\documentclass[preprint,12pt]{elsarticle}

\usepackage{geometry}
\usepackage{amssymb}

\usepackage{colortbl}

\usepackage{slashbox}
\usepackage{amsmath}
\usepackage{bm}
\usepackage{amssymb}
\usepackage{subcaption}
\usepackage{algorithm} 
\usepackage{algorithmic}
\usepackage{flushend}
\usepackage{graphics}
\usepackage{graphicx}
\usepackage{float}
\usepackage{array}
\usepackage{color}
\usepackage{fixltx2e}

\usepackage{multirow}
\usepackage{booktabs}
\usepackage{stfloats}

\usepackage{url}

\usepackage{stfloats}
\usepackage{flushend}
\usepackage{enumerate}
\usepackage{bbding} 

\usepackage{hyperref}					
\hypersetup{colorlinks,
	linkcolor=blue,%
	citecolor=blue}

\usepackage{fancyhdr}
\usepackage[font=small,labelfont=bf,labelsep=none]{caption}
\biboptions{numbers,sort&compress}

\begin{document}

\captionsetup[figure]{labelfont={bf},labelformat={default},labelsep=period,name={Fig.}} 

\begin{frontmatter}

\title{An Adaptive Deep Learning Framework for Day-ahead Forecasting of Photovoltaic Power Generation}

\author[label1]{Xing Luo}
\author[label2,label1]{Dongxiao Zhang\corref{cor1}}
\cortext[cor1]{Corresponding author: Dongxiao Zhang (zhangdx@sustech.edu.cn)}

\address[label1]{Intelligent Energy Lab, Peng Cheng Laboratory, Shenzhen, 518055, P. R. China}
\address[label2]{School of Environmental Science and Engineering, Southern University of Science and Technology, Shenzhen, 518055, P. R. China}


\begin{abstract}
Accurate forecasts of photovoltaic power generation (PVPG) are essential to optimize operations between energy supply and demand. Recently, the propagation of sensors and smart meters has produced an enormous volume of data, which supports the development of data-based PVPG forecasting. Although emerging deep learning (DL) models, such as the long short-term memory (LSTM) model, based on historical data, have provided effective solutions for PVPG forecasting with great successes, these models utilize offline learning. As a result, DL models cannot take advantage of the opportunity to learn from newly-arrived data, and are unable to handle concept drift caused by installing extra PV units and unforeseen PV unit failures. Consequently, to improve day-ahead PVPG forecasting accuracy, as well as eliminate the impacts of concept drift, this paper proposes an adaptive LSTM (AD-LSTM) model, which is a DL framework that can not only acquire general knowledge from historical data, but also dynamically learn specific knowledge from newly-arrived data. A two-phase adaptive learning strategy (TP-ALS) is integrated into AD-LSTM, and a sliding window (SDWIN) algorithm is proposed, to detect concept drift in PV systems. Multiple datasets from PV systems are utilized to assess the feasibility and effectiveness of the proposed approaches. The developed AD-LSTM model demonstrates greater forecasting capability than the offline LSTM model, particularly in the presence of concept drift. Additionally, the proposed AD-LSTM model also achieves superior performance in terms of day-ahead PVPG forecasting compared to other traditional machine learning models and statistical models in the literature.

\end{abstract}


\end{frontmatter}



\newpage
\section{Introduction}
\label{sec:1}

Energy production is a core factor in economic development and technological growth. Indeed, it constitutes energy demand has caused an enormous impact on the global environment, since approximately two-thirds of global greenhouse gas (GHG) is generated from the indispensable basis of rapid urbanization, industrialization, and modernization of a nation \cite{Zhifeng-Guo-2018}. Based on a report from the U.S. Energy Information Administration (EIA), total world energy consumption is expected to increase by 28\% between 2015 and 2040 \cite{Mohammad-Navid-2021}. Meanwhile, the increasing energy production through the massive usage of fossil fuels \cite{Zekai-Sen-2004, C-A-Varotsos-2019}. If this situation continues, worse global warming and climate changes will irreparably result, which are highly correlated with certain meteorological catastrophes, such as violent storms, floods, and droughts.

In response, renewable energy sources have increasingly replaced traditional fossil energy resources in the last decades, among which solar energy is the main source and occupies a high penetration rate in the market \cite{Xing-Luo-2021}. Currently, solar energy technologies, such as photovoltaic (PV), concentrated PV (CPV) and concentrated solar thermal power (CSP), are functionally ready and are almost financially competitive to extract clean and inexhaustible power from the Sun on a large scale \cite{Rich-H-2013}. To fight against climate change, reduce pollution and provide accessible power to people in rural areas, the global capacity of solar power is growing rapidly. The worldwide installed capacity from solar power has risen dramatically from 40 GW in 2010 to 578 GW in 2019, and it is estimated to reach 1700 GW by 2030, based on the International Renewable Energy Agency (IRENA) \cite{M-N-Akhter-2019}.

Solar energy is converted from solar irradiance received by PV panels through the photovoltaic effect \cite{Xing-Luo-2021}. However, surface solar irradiance is highly variable and uncertain due to the complex interaction between radiation and atmospheric constituents, including water vapor, aerosols and clouds, which possess high temporal and spatial variability \cite{Matthew-Lave-2010}. In addition, other local meteorological factors, such as temperature, humidity and wind speed, which also have considerable impacts on the output of PV systems, are variable, as well \cite{Muhammad-Qamar-Raza-2016}. As a result, the PV power generation (PVPG) of a system is highly uncertain and varies dynamically with time. However, power grids need to balance generation and consumption in real-time. Integrating uncertain PV power into existing power systems as a backup electric supply without utilizing any energy reserve devices is not viable. Sudden drops in power production will appear frequently, cause fluctuations in grid voltage, and adversely affect grid stability, potentially even producing grid failures \cite{Alessandro-Ferrara-2021}. Therefore, a reliable PVPG forecast method is urgently needed, and it will markedly decrease the impact of this uncertainty and promote the stability of hybrid power systems.

Accurate PVPG forecasting is a crucial research arena, and a substantial body of research has been devoted to it. In general, PVPG forecasting methodologies illustrated in previous works can be divided into four major categories: the physical model \cite{Alberto-Dolara-2015, Daniel-Koster-2019}; the statistical model \cite{Vagro-2016, John-Boland-2016, X-Zhang-2019, H-Sheng-2018}; the machine learning (ML) model \cite{Liu-Zhao-2016, John-Paul-Mueller-2016, M-Abuella-2015, J-Liu-2015, Y-S-Manjili-2018}; and the hybrid model \cite{Guido-Cervone-2017, Abinet-Tesfaye-Eseye-2018, Fei-Wang-2019}. In recent years, widely-installed smart metering devices have collected an enormous volume of data, which is advantageous to the reformation of ML techniques \cite{Xing-Luo-2021}. As an important branch of ML, deep learning (DL) has attracted substantial attention, and has recently been widely adopted in various engineering fields. DL models possess strong predictive capabilities, in terms of learning feature representations and modeling complex relationships that are commonly hidden in big data \cite{Huaizhi-Wang-2019}. Typical representatives of DL models include the fully-connected neural network (FCNN) \cite{M-Abuella-2015, J-Liu-2015, Changsong-Chen-2011}, the convolutional neural network (CNN) \cite{Fei-Wang-2019}, and the recurrent neural network (RNN) \cite{Massaoudi-2019, Wilms-2018, Kusuma-2021}. Among them, RNN has greatly advanced PVPG forecasting, since time-dependence among sequential data is considered by RNN via its special self-looped structure.

In addition, long short-term memory (LSTM) is a prominent representative in the RNN class, and it has been broadly adopted in time-series regression problems, such as energy forecasting. Compared to the conventional RNN, LSTM can not only handle the correlation of sequential data, but also prevent gradient exploding and gradient vanishing, which are commonly found within conventional RNNs \cite{Xing-Luo-2021}. Consequently, LSTM has achieved great successes in solar energy forecasting in practice \cite{Xing-Luo-2021, Jianqin-Zheng-2020, A-Gensler-2016, Abdel-Nasser-2017}. For example, to solve the problem of multi-region solar power generation forecasting, Zheng et al. \cite{Jianqin-Zheng-2020} proposed the LSTM model incorporated with the particle swarm optimization (PSO) algorithm. The PSO algorithm was adopted to optimize the setting of hyper-parameters within LSTM. Moreover, Gensler et al. \cite{A-Gensler-2016} proposed the LSTM model assisted by an auto-encoder (Auto-LSTM) to solve the problem of solar power forecasting. The results using Auto-LSTM indicate superior performance compared to FCNN, as well as other reference models. Furthermore, Nasser and Mahmoud \cite{Abdel-Nasser-2017} applied the LSTM model to forecast PV power output based on multiple hourly PV datasets collected from different sites. The achieved results also demonstrated superiority of the LSTM model over other compared models.

Despite the remarkable successes achieved in previous works, challenges remain in solar energy forecasting by using DL models. First, current DL models are trained once by periodically going through all data, and then used to infer future PV output \cite{Xing-Luo-2021, Jianqin-Zheng-2020, A-Gensler-2016, Abdel-Nasser-2017}. This approach, however, cannot take advantage of underlying information that could be offered by the newly-arrived data stream. The information hidden in the newly-arrived data stream can have positive impacts on model training and assist the model to obtain better forecasting performances of the near future. Certainly, the model can be re-trained by utilizing the previous data along with the newly-arrived data, but this will incur enormous computational resources with the amount of data increasing. As a consequence, it is not practical in real-life applications. Ideally, DL models should be able to learn from the newly-arrived data immediately, without re-training the previous data. Second, underlying distributions of the streaming data in the energy domain vary over time due to certain unforeseeable changes, causing what is referred to as concept drift \cite{Lu-Jie-2019}. For instance, installing extra PV units and unforeseen PV unit failures of a PV system will inevitably alter data distributions, which results in the collapse of existing mapping relationships between the input variables and the output variable. Current DL models, however, have not accounted for the effects that may be produced by concept drift. In the presence of concept drift, traditional DL models lose efficacy, and forecasting performance will be significantly degraded. Third, offline DL models which are trained by conventional batch learning methods can produce accurate forecasts with minimum errors for the complete testing set, but cannot guarantee the forecasting accuracy of an assigned day, since the model is an averaged one and determined by minimizing the global errors of the testing set. This causes challenges of power system operations by introducing inaccurate day-ahead PV power into grids in practice.

Considering the above limitations in previous investigations, much room exists to improve the performance of DL models. Consequently, this work proposes an adaptive LSTM (AD-LSTM) model, which is a DL framework that can not only acquire general knowledge from historical data, but also dynamically learn specific knowledge from newly-arrived data. It aims to improve day-ahead PVPG forecasting accuracy, as well as eliminate the impacts of concept drift. The main contributions of this work can be summarized as follows:\vspace{-0.13in}

\begin{enumerate}[(1)]
\item Different from conventional offline DL models, the proposed AD-LSTM model not only takes advantage of historical data, but is also capable of continuously learning from newly-arrived data via an integrated two-phase adaptive learning strategy (TP-ALS). \vspace{-0.1in}
\item A sliding window (SDWIN) algorithm is proposed to detect concept drift in PV systems. The AD-LSTM model assisted by SDWIN is better suited for PVPG forecasting than the conventional LSTM, particularly in the presence of concept drift. \vspace{-0.1in}
\item The proposed AD-LSTM model can significantly improve day-ahead PVPG forecasting accuracy to different extents with or without concept drift, compared to representatives of ML and statistics models, to meet the demands of power system operations. \vspace{-0.1in}
\item  A sensitive analysis of AD-LSTM with different model structures is provided. The obtained results verify the feasibility and effectiveness of the proposed AD-LSTM model.
\end{enumerate}

The remainder of this paper is organized as follows. In Section \ref{sec:2}, the basic LSTM model is described. Subsequently, the AD-LSTM model is proposed in Section \ref{sec:3}. The TP-ALS strategy and SDWIN algorithm are illustrated specifically. Afterwards, the data acquisition methods for AD-LSTM are introduced in Section \ref{sec:4}. In Section \ref{sec:5}, several experiments are carried out to assess the proposed approaches. The day-ahead PVPG forecasting performances achieved by different models are compared, as well. Finally, the work is discussed and concluded in Section \ref{sec:6}.

\section{Basic Long Short-term Memory}
\label{sec:2}

\begin{figure}[H]
\centering
\begin{minipage}[t]{0.85\textwidth}\label{fig_6_Drift_Detection_PVPG_a}
\centering
\includegraphics[width=\textwidth]{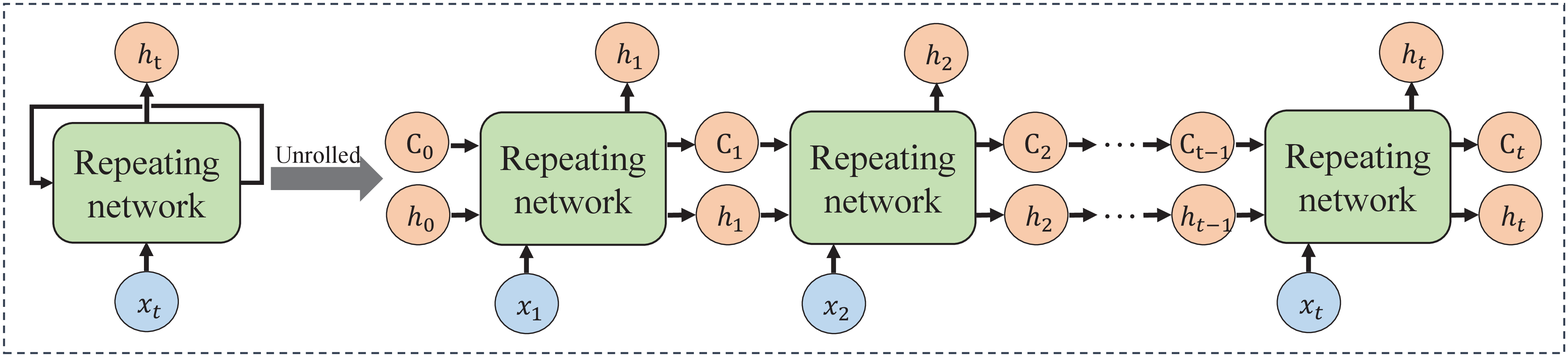}\vspace{-0.3in}
\subcaption{Unrolled self-looped structure of LSTM}
\end{minipage} \\ 
\begin{minipage}[t]{0.55\textwidth}\label{fig_6_Drift_Detection_PVPG_c}
\centering
\includegraphics[width=\textwidth]{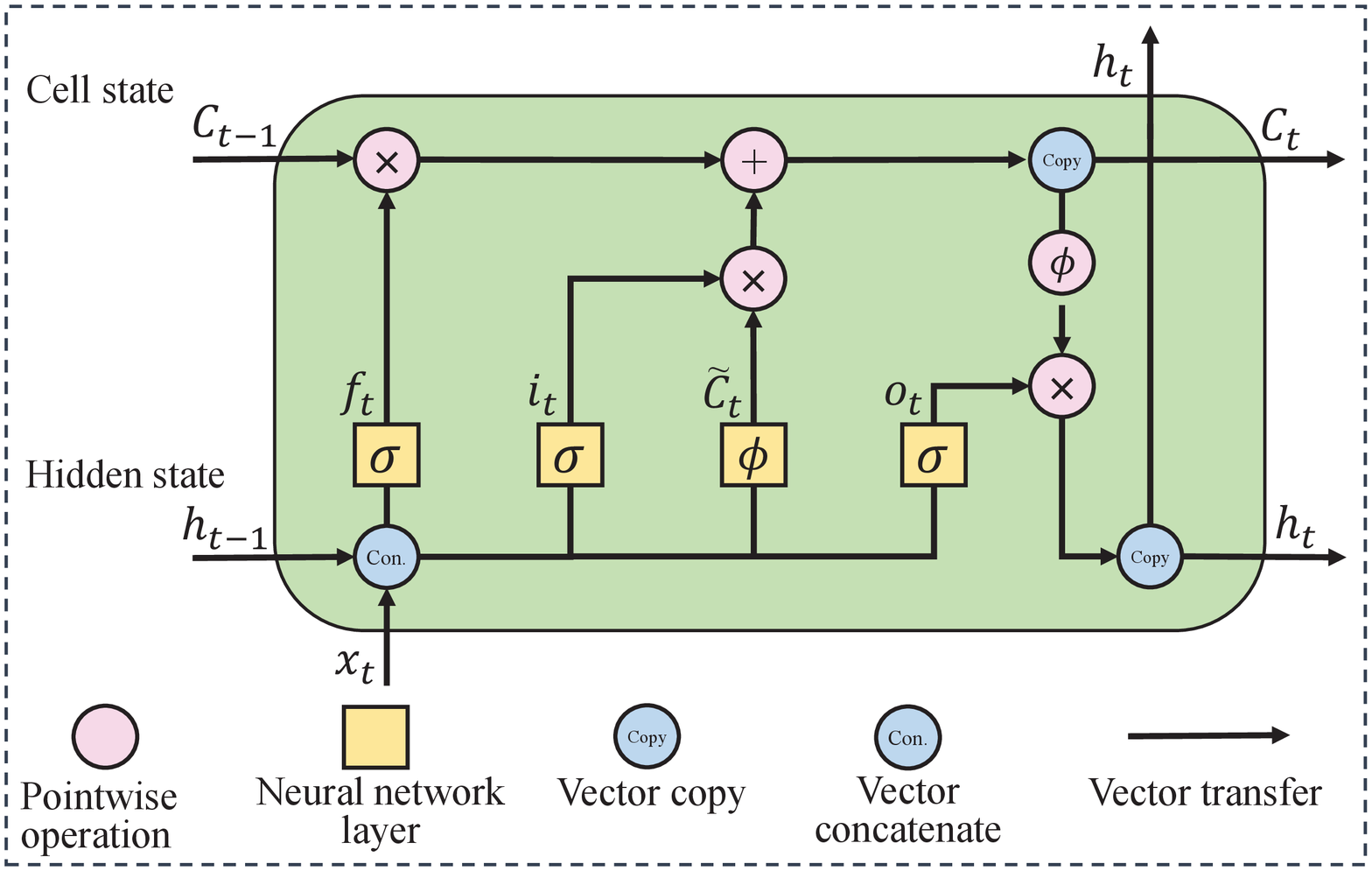}\vspace{-0.3in}
\subcaption{Repeating network of LSTM}
\end{minipage} \\ \vspace{-0.1in}
\caption{Architecture of a typical LSTM model.}
\label{fig:Archi_LSTM_Model}
\end{figure}

As discussed in Section \ref{sec:1}, LSTM outperforms other conventional DL models in solving sequential-data based regression problems, due to the existence of an internal self-looped structure, as shown in Fig. \ref{fig:Archi_LSTM_Model} (a). In view of the wide applications and successes of LSTM in energy forecasting \cite{M-Schuster-1997, Jianqin-Zheng-2020, A-Gensler-2016, Abdel-Nasser-2017}, LSTM is taken as an appropriate DL framework to incorporate with the proposed adaptive strategy in this work.

%

As a special kind of recurrent neural network (RNN), LSTM contains four well-designed layers, i.e., the ``forget gate" layer, the ``input gate" layer, the ``output gate" layer, and the \textit{Tanh} layer, as illustrated in Fig. \ref{fig:Archi_LSTM_Model} (b). In an LSTM unit, as a core variable, the cell state, $C_{t}$, can carry and transmit information between different time steps. The gate layer structure carefully regulates what information can be added or removed to the cell state. By using an activation function, which is usually the sigmoid function expressed in Eq. (\ref{eq.Sigmoid_Function}), these three gate layers all output a coefficient value between zero and one.
\begin{equation}\label{eq.Sigmoid_Function}
\sigma(x) = \frac{1}{1 + e^{-x}}
\end{equation}

Specifically, the LSTM unit determines what stored information should be discarded from the previous cell state, $c_{t-1}$, by the first layer, i.e., the ``forget gate" layer. The output of the forget gate, $f_{t}$, can be mathematically illustrated as follows:
\begin{equation}\label{eq.Forget_Gate_Layer}
f_{t} = \sigma (W_{f} \cdot [h_{t-1}, x_{t}] + b_{f} )
\end{equation}

Afterwards, the LSTM unit determines what information can be stored in the new cell state. The process comprises two parts, which are regulated by the second layer, termed the ``input gate" layer, and the third layer, termed the \textit{Tanh} layer, respectively. To decide the proportion of the new information, the input gate, $i_{t}$, also outputs a coefficient value between zero and one. Meanwhile, the \textit{Tanh} layer, which is designed based on \textit{Tanh} function (illustrated in Eq. (\ref{eq.Tanh_Function})), is utilized to produce new candidate values as a vector that can be stored in the state. The output of the \textit{Tanh} layer, $\tilde{C}_{t}$, is a number between -1 and 1. Therefore, the outputs of the input gate layer and \textit{Tanh} layer can be mathematically illustrated in Eqs. (\ref{eq.Input_Gate_Layer}) and (\ref{eq.Tanh_Layer}), respectively:
\begin{equation}\label{eq.Tanh_Function}
Tanh(x) = \frac{e^{x}-e^{-x}}{e^{x}+e^{-x}}
\end{equation}

\begin{equation}\label{eq.Input_Gate_Layer}
i_{t} = \sigma (W_{i} \cdot [h_{t-1}, x_{t}] + b_{i} )
\end{equation}

\begin{equation}\label{eq.Tanh_Layer}
\tilde{C}_{t} = \phi (W_{C} \cdot [h_{t-1}, x_{t}] + b_{C} )
\end{equation}

After the first three layers, the old cell state, $C_{t-1}$, is updated into the new cell state, $C_{t}$. In previous layers, the forget gate, $f_{t}$, determines what information has to be dropped, and the input gate, $i_{t}$, determines what information can be added to the new cell state, $\tilde{C}_{t}$. Then, the updating process can be mathematically expressed in Eq. (\ref{eq.Updating_Process}):
\begin{equation}\label{eq.Updating_Process}
C_{t} = f_{t} \ast C_{t-1} + i_{t} \ast \tilde{C}_{t}
\end{equation}

Finally, the ``output gate" layer generates the final output of the LSTM unit, $o_{t}$, based on the updated cell state, $C_{t}$. This process is presented in Eq. (\ref{eq.Output_Layer}):
\begin{equation}\label{eq.Output_Layer}
o_{t} = \sigma (W_{o} \cdot [h_{t-1}, x_{t}] + b_{o} ) \ast \phi (C_{t})
\end{equation}

In Eqs. (\ref{eq.Forget_Gate_Layer}) and (\ref{eq.Input_Gate_Layer}) - (\ref{eq.Output_Layer}), parameter vectors, $W = [W_{f}, W_{i}, W_{C}, W_{o}]$ and $b = [b_{f}, b_{i}, b_{C}, b_{o}]$, represent the weights and bias coefficients of each layer, respectively. Let $\theta = \{W, b\}$, the loss function of LSTM, $\mathcal{L}(\theta)_{\text{LSTM}}$, which is set as the mean square error (MSE) between the ground truth data and the model outputs, can then be formulated as follows:
\begin{equation}\label{eq.Loss_Function}
\mathcal{L} (\theta)_{\text{LSTM}} = \frac{1}{N}\sum^{N}_{i=1} \vert \hat{y}_{i} - y_{i} \vert ^{2}
\end{equation}
where $\hat{y}_{i} = \text{NN}(x_{i};\theta)$ is the $i^{\text{th}}$ forecast of LSTM; and $N$ is the total number of observed data. In the training process of LSTM, the parameter vector of the network, $\theta$, will be tuned continuously by minimizing the loss function via an optimization algorithm, such as adaptive moment estimation (Adam) or stochastic gradient descent (SGD) \cite{Xing-Luo-2021}. In this way, the LSTM model can be well-trained. However, current DL models employ offline learning without considering the underlying information that could be provided by the newly-arrived data. Therefore, an adaptive LSTM model is proposed in Section \ref{sec:3}.

\section{Proposed Adaptive Long Short-term Memory Model}
\label{sec:3}

\begin{figure}[H]
  \centering
  \includegraphics[width=9.5 cm]{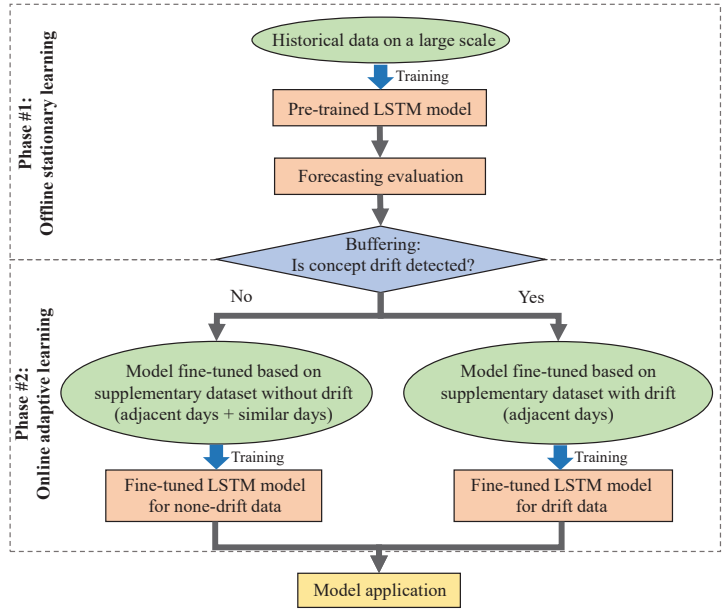}\vspace{-0.1in}
  \caption{Flow chart of the integrated two-phase adaptive learning strategy for AD-LSTM.}
  \label{fig:Adaptive_Learning_Strategy}
\end{figure}

In this section, we propose an adaptive LSTM (AD-LSTM) model, which is a PVPG forecasting framework that can not only acquire general knowledge from historical data, but also dynamically learn specific knowledge from newly-arrived data. Different from conventional DL models in previous investigations, which are normally offline, the proposed AD-LSTM is the combination of the basic LSTM and an integrated two-phase adaptive learning strategy (TP-ALS). According to the flow chart of TP-ALS, as shown in Fig. \ref{fig:Adaptive_Learning_Strategy}, the learning process of AD-LSTM consists of two connected phases, termed the offline stationary learning phase and the online adaptive learning phase, respectively.

Specifically, in the first phase, the basic LSTM model is trained on a large amount of historical data first and then evaluated based on an individual testing set, as done by most conventional learning methods in previous works. By doing so, the LSTM model can learn general characteristics of PVPG variations, and establish robust mapping relationships covering a variety of patterns between the input space and the output space.

\begin{figure}[!tp]
  \centering
  \includegraphics[width=12 cm]{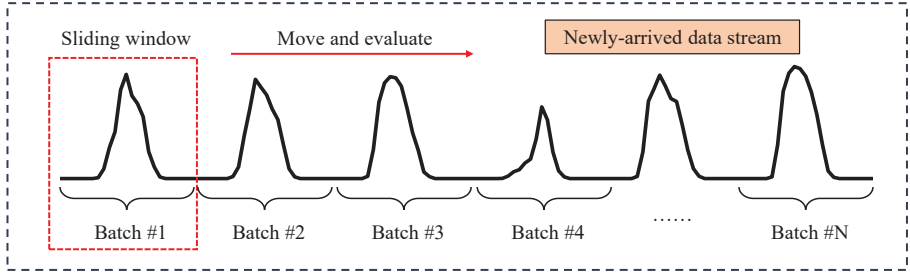}\vspace{-0.1in}
  \caption{Buffering module with a sliding window for concept drift detection.}
  \label{fig:Sliding_Window}
\end{figure}

Subsequently, prior to the second phase, a crucial task is to detect if concept drift appears in the newly-arrived data. To achieve this, a buffering module with a sliding window (SDWIN) algorithm is proposed, as illustrated in Fig. \ref{fig:Sliding_Window}. The main purpose of the buffering module is to detect and temporally store batches, in which the pre-trained model performs very poorly. In general, the well-trained DL model can produce accurate forecasts for data without drift. However, in the presence of concept drift, the pre-trained model is no longer suited to the newly-arrived data, and its predictive performance will be significantly diminished since the data distributions have already been changed. To address this, the embedded SDWIN algorithm utilizes a sliding window to continuously evaluate the forecasting errors of batches within the streaming data. A significant rise of batch error suggests a drift in this window. The SDWIN algorithm attempts to discern the occurrence of concept drift by continuously monitoring the error variation. Mean square error (MSE) is adopted as the evaluation criterion, and the MSE of $j^{\text{th}}$ batch can be mathematically expressed as follows:
\begin{equation}\label{eq.Batch MSE}
E_{j} = \frac{1}{N_{b}}\sum^{N_{b}}_{i=1} \vert \hat{y}_{i} - y_{i} \vert ^{2}
\end{equation}
where $\hat{y}_{i}$ and $y_{i}$ are the forecasted and observed values, respectively; and $N_{b}$ is the total number of the ground truth data within a batch. Since we are concerned with day-ahead forecasting, one batch includes the data of 24 hours. Defining an amount of streaming data consisting of $N$ batches, a pre-trained LSTM model as $M_{\text{pt}}$, an error threshold $E_\text{th} = \bar{E} + 3 E_\text{std}$, where $\bar{E}$ and $E_\text{std}$ are the mean and standard deviation of errors of a period of time, and a confidential level as $C$ and its maximum value as $C_\text{max}$, the SDWIN method can be illustrated in Algorithm \ref{agr:SDWIN}.

\begin{algorithm}[!htbp]
    \caption{SDWIN algorithm.}
    \begin{algorithmic}[1]
    \label{agr:SDWIN}
\REQUIRE Pre-trained model $M_{\text{pt}}$, streaming data including $N$ batches.
\ENSURE Determination of concept drift detection.
\STATE \textbf{FOR} $j = 1 \rightarrow N$, evaluate Batch \#$j$:
\STATE \ \ \ \ \textbf{IF} $E_{j} > E_\text{th}$:
\STATE \ \ \ \ Batch \#$j$ $\leftarrow$ $\text{warning}$, window slides to the next batch.
\STATE \ \ \ \ Confidential level of drift $\leftarrow$ improved, $C = C + 1$.
\STATE \ \ \ \ \ \ \ \ \textbf{IF} $C \geqslant C_{\text{max}}$:
\STATE \ \ \ \ \ \ \ \ Concept drift detected, break \textbf{FOR} loop.
\STATE \ \ \ \ \ \ \ \ \textbf{END IF}
\STATE \ \ \ \ \textbf{ELSE}
\STATE \ \ \ \ Warnings clear, window slides to the next batch.
\STATE \ \ \ \ Confidential level reset, $C \leftarrow 0$.
\STATE \ \ \ \ \textbf{END IF}
\STATE \textbf{END FOR}
\end{algorithmic}
\end{algorithm}

In the second phase, the pre-trained model is transferred and fine-tuned based on a supplementary dataset consisting of the training data that may have similar patterns to the target day. Theoretically, using highly correlated data as the supplementary dataset is beneficial for model training and improving the forecasting accuracy of the target day. As a result, two types of data sources are considered in the supplementary dataset, termed ``adjacent days" (AD) and ``similar days" (SD). The AD sub-dataset consists of the newly-arrived data several days prior to the target day; whereas, the SD sub-dataset consists of the highly correlated data, which have similar weather conditions to the target day in history.

Depending on whether or not concept drift is detected, two possible situations are taken into account and discussed. (\romannumeral1) Concept drift is not detected. Then the supplementary dataset contains both AD and SD sub-datasets as a consequence. On the one hand, the information underlying the newly-arrived data (e.g., seasonal characteristics) may have positive impacts on the model training. On the other hand, the LSTM model using the highly correlated data in history as a part of the training dataset is much better than the overall model due to the advantage of eliminating the unknown interference caused by other meteorological factors. In this way, it can enhance the capability of LSTM in terms of forecasting near future days. (\romannumeral2) Concept drift is detected. The supplementary dataset consequently only contains the AD sub-dataset. The SD sub-dataset from historical data is no longer suited for the model training, as the data distributions have apparently been changed in this case. Using the AD sub-dataset as the supplementary dataset for fine-tuning allows the model to adjust itself rapidly to capturing the new patterns under streaming data. It is worth noting that when concept drift is not confirmed during the buffering time, the newly-arriving data are treated as none-drift data.

Essentially, pre-training the DL model based on a large amount of historical data in the first phase of TP-ALS can improve the generalization ability of the DL model. As a result, the well-trained model can provide PVPG forecasts with acceptable accuracy if the data follow the same distributions. Based on this, the pre-trained model can be used in the SDWIN algorithm to identify the presence of concept drift. In addition, in the second phase of TP-ALS, fine-tuning the DL model based on a supplementary dataset that contains highly correlated data to the target day can considerably strengthen the forecasting capability of the DL model to the target day, thus achieving the goal of improving day-ahead PVPG forecasting accuracy. 

\section{Data Acquisition Methods for the AD-LSTM Model}
\label{sec:4}

In this section, we introduce the methods used to obtain the supplementary dataset for the AD-LSTM model. As discussed in Section \ref{sec:3}, two types of data sources, i.e., AD and SD, are required for model fine-tuning. Specifically, it is not difficult to obtain an AD sub-dataset by deriving it from the newly-arrived data stream. Data of several days prior to the target day can be easily extracted. The size of AD and SD sub-datasets will be analyzed and discussed in Subsection \ref{sec:53}.

Subsequently, in addition to the AD sub-dataset, a weighted K-nearest neighbor (W-KNN) is proposed to acquire the SD sub-dataset. In statistics, the basic K-nearest neighbor (KNN) is a non-parametric method, and can be used for both classification and regression \cite{Yao-Zhang-2016}. In this work, KNN is utilized to create the SD sub-dataset by extracting $K$ historical samples which have similar meteorological conditions to the target day. Firstly, to assess the similarity between any two typical days, an evaluation measure is required. In this work, Euclidean distance is adopted and enhanced by using weighting as follows:
\begin{equation}\label{eq.Euclidean_Distance}
D(X^{\text{targ}}, X^{\text{hist}}) = \sqrt{\sum\limits_{i = 1}^{N_{v}} w_{i} \cdot (x^{\text{targ}}_{i} - x^{\text{hist}}_{i})^{2} }
\end{equation}
where $X^{\text{targ}}$ and $X^{\text{hist}}$ are two compared metrics expressing all input variables, i.e., meteorological variables of the target day and the sample day in history, respectively; $x^{\text{targ}}_{i}$ and $x^{\text{hist}}_{i}$ are vectors denoting the $i^{\text{th}}$ input variable over a typical day in $X^{\text{targ}}$ and $X^{\text{hist}}$, respectively; $N_{v}$ represents the total number of input variables; and $w_{i}$ is the corresponding weight assigned to the $i^{\text{th}}$ input variable. In the basic KNN, the weight of an individual input variable is equal, which indicates that each input variable makes an equal contribution to the calculated distance $D$. Nonetheless, the significance of different variables to the PVPG is apparently dissimilar. For example, solar radiation is the most significant input variable for the determination of the final PV output. Therefore, it deserves a higher value of weight compared to other input variables when calculating similarity.

Consequently, we propose W-KNN, in which the weight, $w_{i}$, denotes the contribution made by each input variable to the final distance. It is more reasonable than the basic KNN. In W-KNN, the weights are determined according to the degree of relevance between the input variable and the target variable by using the Pearson correlation coefficient (PCC), which is a measure of the linear correlation between variables $\mathcal{X}$ and $\mathcal{Y}$ in statistics. By doing so, the input variable which has a higher correlation with the target variable can have a higher value of weight in distance calculation. PCC can be mathematically defined as follows:

\begin{equation}\label{eq.PCC}
\rho(\mathcal{X}, \mathcal{Y})=\frac{\sum\limits_{j=1}^{N_{s}} (x_{j}-\overline{x}) \cdot (y_{j}-\overline{y})} {\sqrt{\sum\limits_{j=1}^{N_{s}} (x_{j}-\overline{x})^{2}} \cdot \sqrt{\sum\limits_{j=1}^{N_{s}} (y_{j}-\overline{y})^{2}} }
\end{equation}
where $x_{j}$ and $y_{j}$ are instances in variables $\mathcal{X}$ and $\mathcal{Y}$, respectively; $\bar{x}=\sum_{j=1}^{N_{s}} x_{j}/N_{s}$ and analogously for $\bar{y}$; and $N_{s}$ is the total number of instances in each variable. It is important to note that the absolute value of the calculated PCC is assigned to the corresponding weight $w_{i}$ in this work. Based on the rank of calculated similarity by W-KNN, $K$ nearest neighbors representing $K$ sample days that have the highest correlations with the target day can be selected and used to form the SD sub-dataset. Together with the AD sub-dataset, the supplementary dataset is obtained.

\section{Results and Discussion}
\label{sec:5}

In Section \ref{sec:5}, experiments are conducted to assess the feasibility and effectiveness of the proposed approaches. The day-ahead PVPG forecasting performances achieved by different models are compared, and numerical results are presented.

\subsection{Data Pre-processing}
\label{sec:51}

\subsubsection{Data description}
A real-life dataset \cite{Tao-Hong-2016} consisting of data records over 27 months from a PV plant located in Australia is utilized to evaluate the proposed methods in this work. The input feature dataset contains weather forecasts of four commonly-used meteorological variables which are obtained from the European Center for Medium-range Weather Forecasts (ECMWF) \cite{Tao-Hong-2016}. The adopted input meteorological variables include: (\romannumeral1) Relative humidity at 1000 mbar (units: \%); (\romannumeral2) 2-meter temperature (units: K); (\romannumeral3) Hourly solar radiation at the land surface (units: $\text{J} \cdot \text{m}^{-2}$); and (\romannumeral4) Hourly net solar radiation at the top of the atmosphere (units: $\text{J} \cdot \text{m}^{-2}$). The output variable is the normalized PVPG. To simulate real situations, data of 15 months are used as the historical data to pre-process DL models as required in TP-ALS; whereas, the remaining data (12 months) are regarded as the newly-arrived data to assess the proposed AD-LSTM model. The horizon of PVPG forecasting is set as one-day-ahead since the aim of this work is to improve day-ahead forecasting accuracy as much as possible. In the meantime, accurate one-day-ahead weather forecasts can be easily obtained in practice.

\subsubsection{Data normalization}
Data normalization is a standard procedure in DL for processing feature data in various ranges into a consistent range with the objective of reducing large feature dominance and improving convergence \cite{Mohammad-Navid-2021}. The min-max normalization (MMN) technique, which transforms the original value, $x_{\text{orig}}$, to the normalized value, $x_{\text{norm}}$, is utilized in this study. The formulation of the MMN is illustrated in Eq. (\ref{eq:Min_Max_Normalization}):
\begin{equation}\label{eq:Min_Max_Normalization}
x_{\text{norm}} = \frac{x_{\text{orig}} - x_{\text{min}}}{x_{\text{max}} - x_{\text{min}}}
\end{equation}
where $x_{\text{min}}$ and $x_{\text{max}}$ are the minimum and maximum values within the dataset, respectively. The MMN technique restricts the data within the range between zero and one.

In this work, the normalization for the AD-LSTM model includes two independent processes, i.e., normalization for the historical data in the offline stationary learning phase and normalization for the newly-arrived data in the online adaptive learning phase (corresponding to TP-ALS in Section \ref{sec:3}). In the former process, all data are available prior to the start of training, and thus normalization for the historical data can proceed by using minimum and maximum values (denoted as $x_{\text{min}}^{\text{hist}}$ and $x_{\text{max}}^{\text{hist}}$, respectively ) within the historical dataset. In the latter process, the complete newly-arrived dataset cannot be determined instantly, although the newly-arrived data have to be processed as they arrive. Consequently, normalization for the newly-arrived data also uses $x_{\text{min}}^{\text{hist}}$ and $x_{\text{max}}^{\text{hist}}$ as the minimum and maximum values, respectively. It is worth noting that all feature data should be normalized prior to transmitting them to DL models.

\subsubsection{Evaluation metrics}
To evaluate the forecasting performances of models from different perspectives, several evaluation metrics, including mean absolute error (MAE), mean square error (MSE), and $\text{R}^{2}$ score, are proposed. Formulations of MAE, MSE, and $\text{R}^{2}$ score can be mathematically illustrated in Eqs. (\ref{eq:MAE}) - (\ref{eq:R2}), respectively:
\begin{equation}\label{eq:MAE}
\text{MAE} = \frac{1}{N} \sum_{i=1}^{N} \vert \hat{y}_{i} - y_{i} \vert
\end{equation}

\begin{equation}\label{eq:MSE}
\text{MSE} = \frac{1}{N} \sum_{i=1}^{N} ( \hat{y}_{i} - y_{i} )^2
\end{equation}

\begin{equation}\label{eq:R2}
\text{R}^{2} = 1 - \frac{\sum\limits_{i=1}^{N} (\hat{y}_{i} - y_{i})^2 } {\sum\limits_{i=1}^{N} (\overline{y} - y_{i})^2 }
\end{equation}
where $\hat{y}_{i}$, $y_{i}$, and $\overline{y}$ represent the forecasted value, the observed value, and the mean of observed values, respectively. 

\subsection{Model Pre-training Based on Historical Data}
\label{sec:52}

According to TP-ALS proposed in Section \ref{sec:3}, the AD-LSTM model has to be pre-trained based on a large volume of historical data at first. The pre-training process assists the model to learn general knowledge from historical data, thus establishing robust mapping relationships covering a variety of patterns between the input variables and the output variable. To cross-validate the forecasting capability of the AD-LSTM model, the data of 12 months are set as the training set, and the data of 3 months are used as the testing set.

\begin{figure}[H]
  \centering
  \includegraphics[width=9.5 cm]{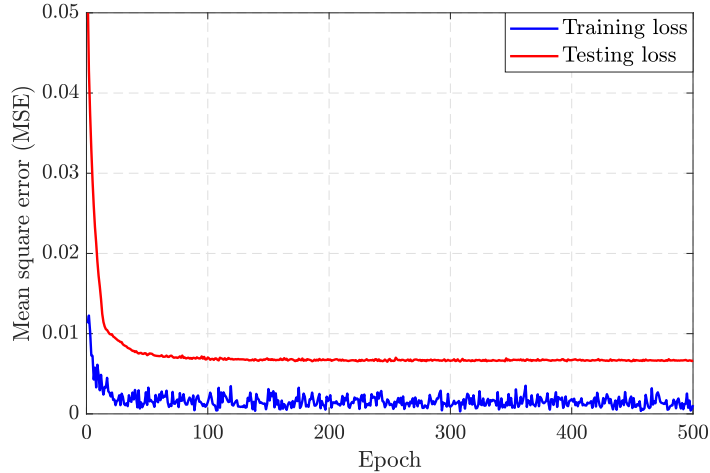}\vspace{-0.1in}
  \caption{Variations of training and testing errors for AD-LSTM in the pre-training process.}
  \label{fig:Loss_Variation}
\end{figure} \vspace{-0.25in}

\begin{table}[H] 
\renewcommand\arraystretch{1.1}
\setlength{\abovecaptionskip}{0pt}
\setlength{\belowcaptionskip}{10pt}
\caption{Illustration of adopted hyper-parameters of AD-LSTM in the pre-training process.}
\centering
\label{tab:Hyper-parameters_Pretrained_Model}\scalebox{0.8}{
\begin{tabular}{lcccccc}
\hline
Hyper-parameter & LSTM layer number & Time step & Hidden unit & Input size & Learning rate & Batch size \\ \hline
Setting value   & 1                 & 4         & 4           & 4          & 0.001         & 256        \\ \hline
\end{tabular}}
\end{table}

In DL, the hyper-parameter setting of a model causes appreciable impacts on the model performance. Therefore, after training the AD-LSTM model based on historical data multiple times, suitable hyper-parameter settings are determined, as shown in Table \ref{tab:Hyper-parameters_Pretrained_Model}. The initial net parameters are set by default, and the Adam optimizer is adopted in the network optimization. The AD-LSTM model is trained persistently to minimize the sum of MSE between forecasts and observations. Variations of training and testing errors for the AD-LSTM model during 500 epochs in the pre-training process are illustrated in Fig. \ref{fig:Loss_Variation}. The model with the lowest testing errors will be saved and utilized to incorporate with the newly-arrived data for improving the day-ahead PVPG forecasting performance. It is important to note that other settings of model structure (e.g., increasing the number of hidden units) may also obtain similar results. However, as in the second phase of TP-ALS, since the well pre-trained model is fine-tuned based on the supplementary dataset on a small scale, it is not necessary to utilize a DL model in practice that is too complex. Overall, the more complicated is the adopted model, the more computational resources are required. A sensitive analysis of AD-LSTM with different structures is provided in the Appendix.

\subsection{Evaluation of Concept Drift Detection}
\label{sec:53}

\begin{figure}[H]
\centering
\begin{minipage}[t]{0.92\textwidth}\label{fig_6_Drift_Detection_PVPG_a}
\centering
\includegraphics[width=\textwidth]{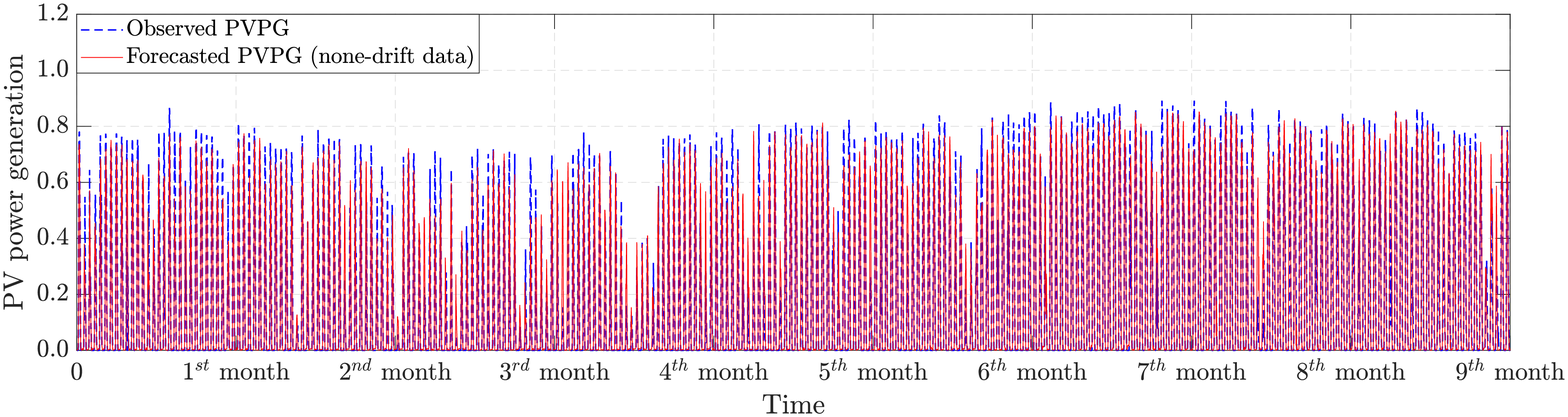}\vspace{-0.3in}
\end{minipage} \\ \vspace{0.3in}
\begin{minipage}[t]{0.92\textwidth}\label{fig_6_Drift_Detection_MSE_b}
\centering
\includegraphics[width=\textwidth]{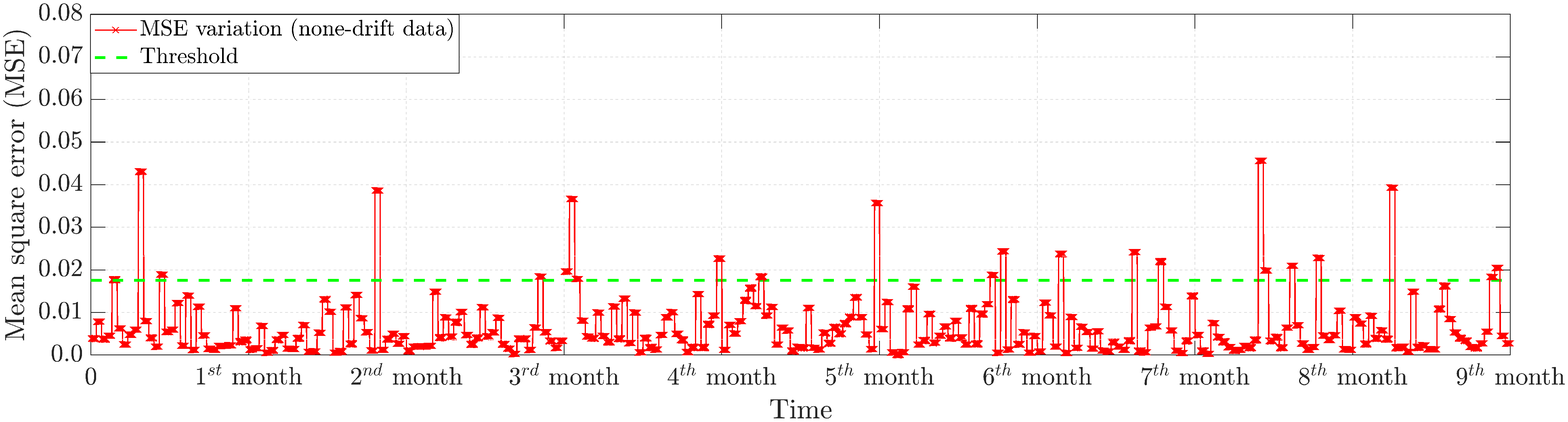}\vspace{-0.35in}
\subcaption{Evaluated based on none-drift data}
\end{minipage}\\ \vspace{0.1in}
\begin{minipage}[t]{0.92\textwidth}\label{fig_6_Drift_Detection_PVPG_c}
\centering
\includegraphics[width=\textwidth]{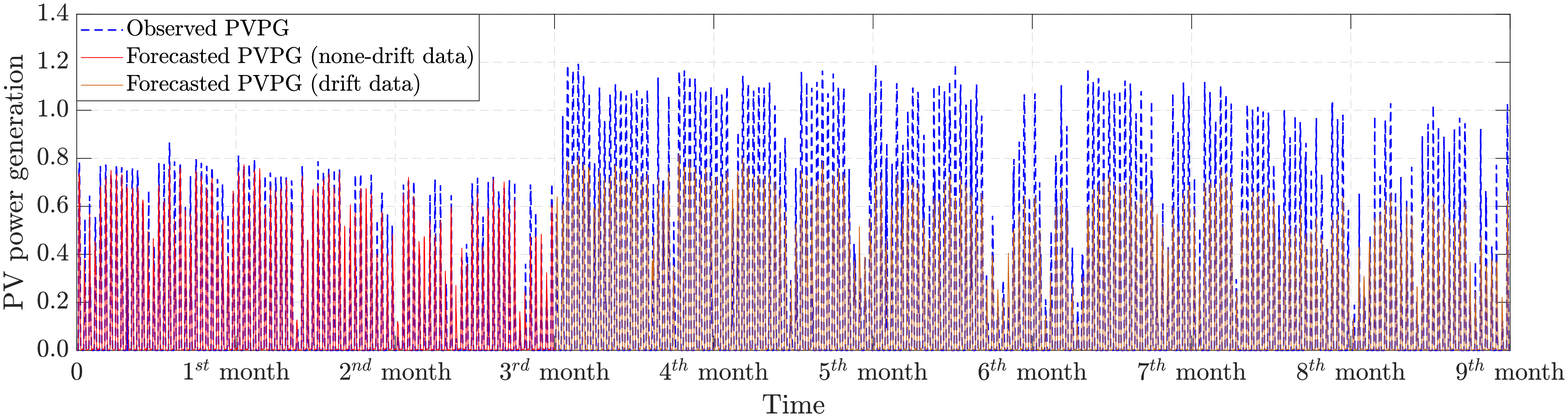}\vspace{-0.3in}
\end{minipage} \\ \vspace{0.3in}
\begin{minipage}[t]{0.92\textwidth}\label{fig_6_Drift_Detection_MSE_d}
\centering
\includegraphics[width=\textwidth]{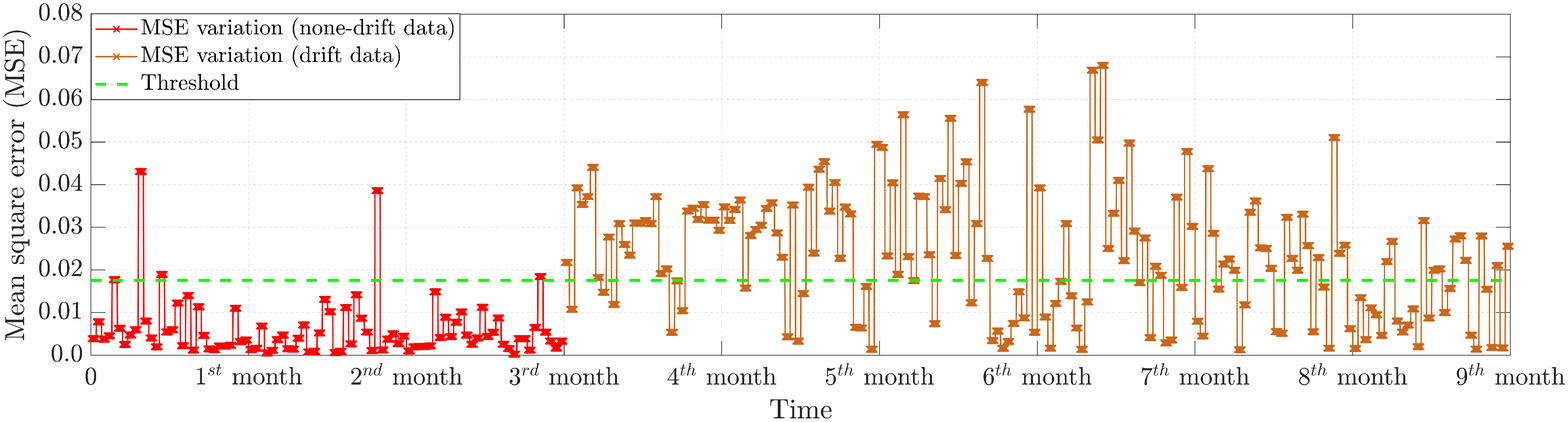}\vspace{-0.35in}
\subcaption{Evaluated based on drift data}
\end{minipage}\\ \vspace{-0.1in}
\caption{Concept drift analysis in PVPG forecasting for both none-drift and drift data.}
\label{fig:Concept_Drift_Detection}
\end{figure}

The major advantage of the AD-LSTM model in energy forecasting is its ability to identify changes in data patterns via the SDWIN algorithm. Therefore, a requisite task prior to PVPG forecasting is to evaluate the occurrence of concept drift within streaming data. In this work, as it is infeasible to obtain drift data from PV plants in practice, real-life data with a bias coefficient are utilized as drift data to simulate situations, such as installing extra PV units and unforeseen PV unit failures of a PV system. Subsequently, the concept drift evaluation for both none-drift and drift data can be conducted, and the corresponding results are presented in Fig. \ref{fig:Concept_Drift_Detection}.

Specifically, the upper figures in Fig. \ref{fig:Concept_Drift_Detection} (a) and (b) illustrate the PVPG forecasting results based on the pre-trained AD-LSTM model, while the related MSE variations are presented in the lower figures. On the one hand, the PVPG forecasts produced by the pre-trained AD-LSTM model based on the none-drift data follow the variations of the observations closely, as shown in the upper figure of Fig. \ref{fig:Concept_Drift_Detection} (a). The results demonstrate the strong forecasting capability of the pre-trained AD-LSTM model, which is developed based on a large amount of historical data. In addition, from the lower figure of Fig. \ref{fig:Concept_Drift_Detection} (a), the corresponding MSE for the none-drift data varies regularly with time. The averaged MSE over 9 months' record is $6.73 \times 10^{-3}$, which is very low in practice. Furthermore, most MSE values are lower than an error threshold as defined in Section \ref{sec:3}, which indicates the strong robustness of the pre-trained AD-LSTM model. Nonetheless, it can also be observed that a few MSE values of PVPG forecasts are higher than the threshold due to the utilization of NWP data. Inaccurate weather forecasts may result in certain difficulties in energy forecasting by using DL models.

On the other hand, the forecasting capability of the model is significantly degraded when the data patterns change, as illustrated in Fig. \ref{fig:Concept_Drift_Detection} (b). The obtained results indicate that the current model is no longer suited for the drift data, and fine-tuning of the model is highly necessary. The MSE variation in the lower figure of Fig. \ref{fig:Concept_Drift_Detection} (b) also reflects such degradations. Expectedly, the averaged MSE greatly increases to $23.41 \times 10^{-3}$, which is over three times higher than the MSE of the none-drift data. In addition, a large number of MSE values exceed the error threshold, indicating occurrences of concept drift with a high probability. In this case, concept drift within newly-arrived data can be detected and confirmed based on the proposed SDWIN algorithm in Section \ref{sec:3}, where the maximum confidential level $C_{\text{max}} = 3$. Indeed, the evaluation results by SDWIN are in accordance with this fact, thus verifying the effectiveness and feasibility of the proposed SDWIW algorithm. After the evaluation of concept drift detection, the newly-arrived data can be used for the fine-tuning of AD-LSTM accordingly.

\subsection{Comparisons Between OL-LSTM and Proposed AD-LSTM}
\label{sec:54}

\begin{figure}[H]
\centering
\begin{minipage}[t]{0.48\textwidth}\label{fig_7_Result_NoDrift_Winter_a}
\centering
\includegraphics[width=\textwidth]{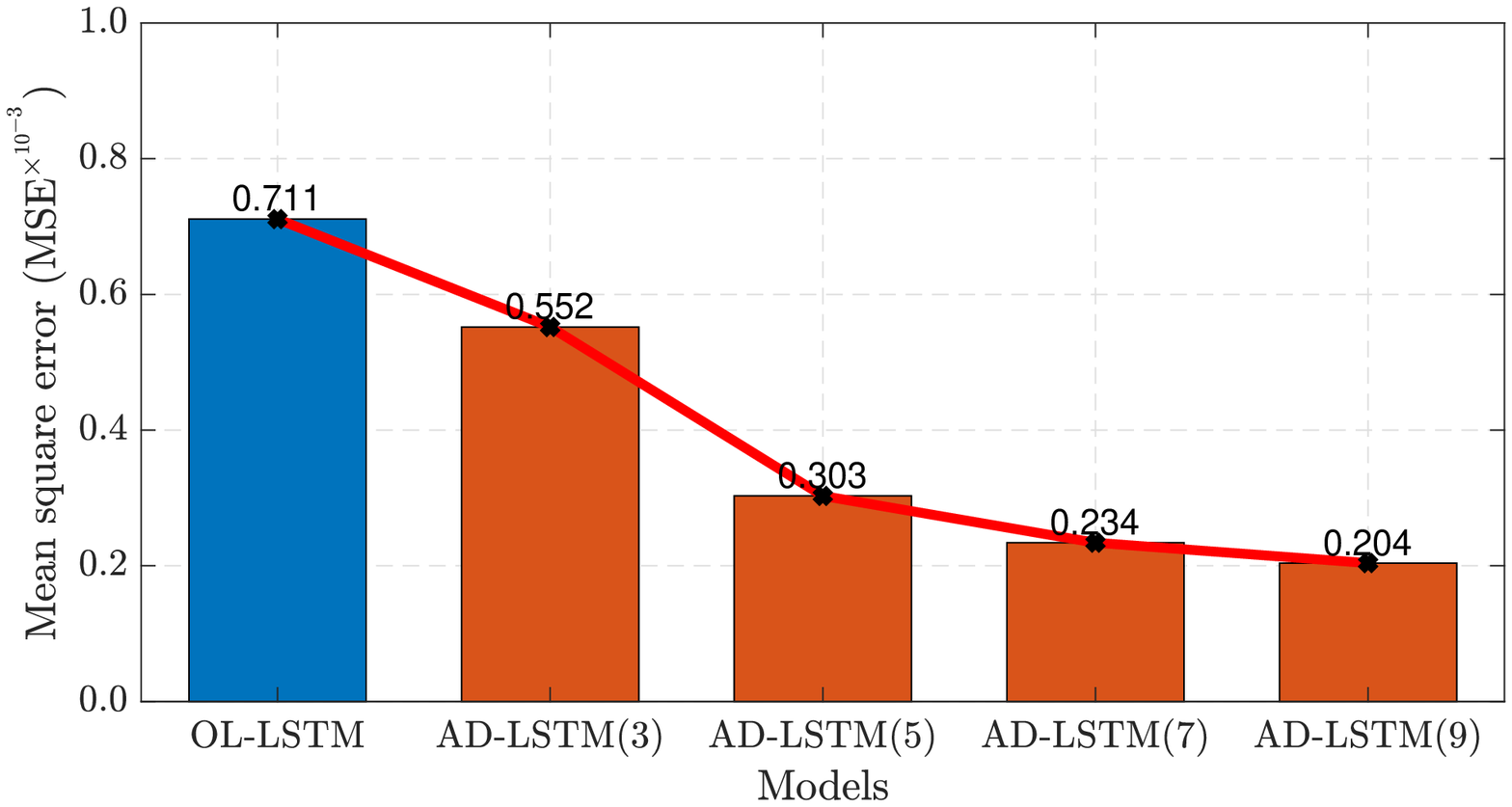}\vspace{-0.35in}
\subcaption{Case \#1 (Winter)}
\end{minipage} 
\begin{minipage}[t]{0.48\textwidth}\label{fig_7_Result_NoDrift_Spring_b}
\centering
\includegraphics[width=\textwidth]{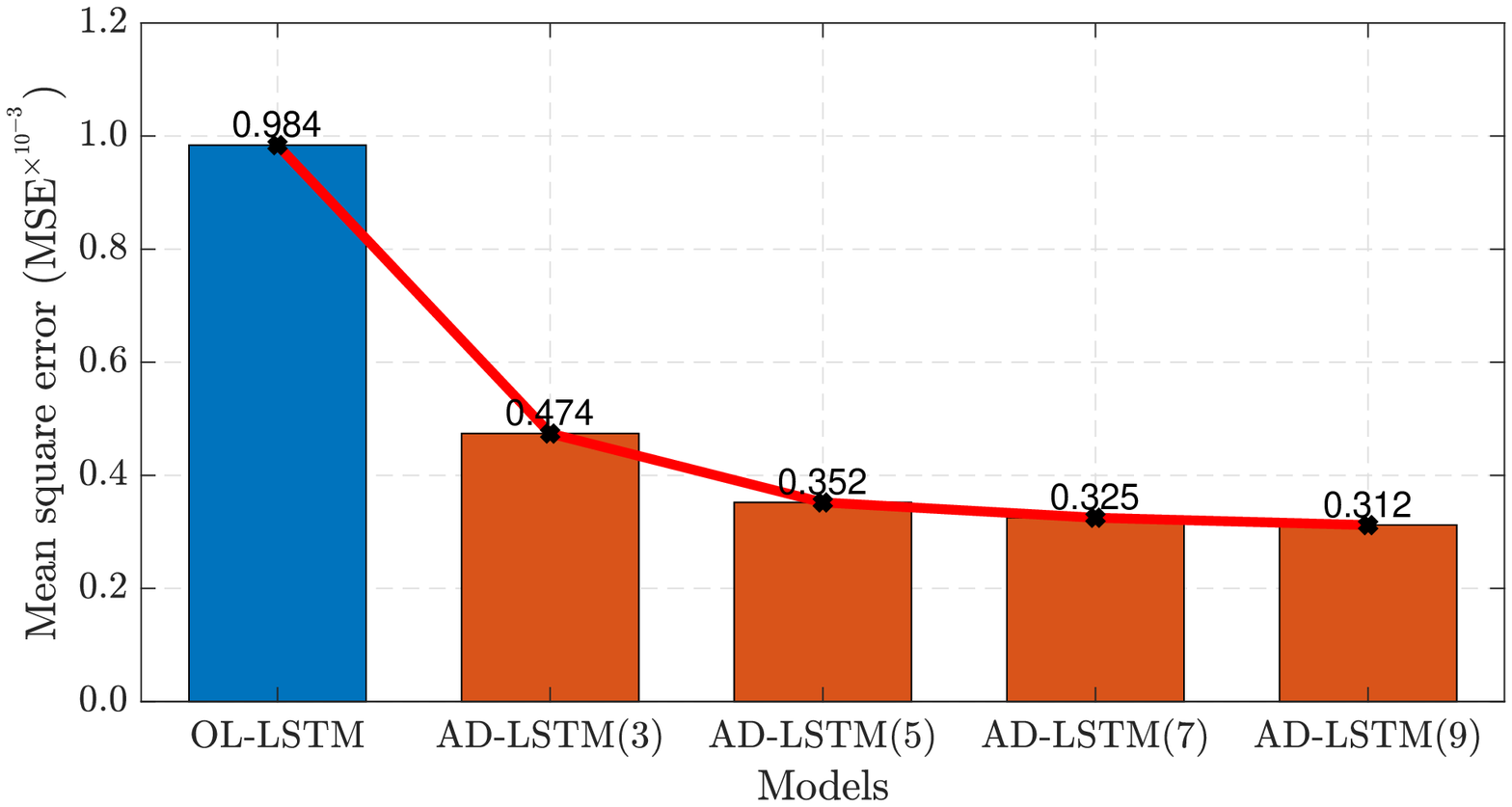}\vspace{-0.35in}
\subcaption{Case \#2 (Spring)}
\end{minipage}\\ 
\begin{minipage}[t]{0.48\textwidth}\label{fig_7_Result_NoDrift_Summer_c}
\centering
\includegraphics[width=\textwidth]{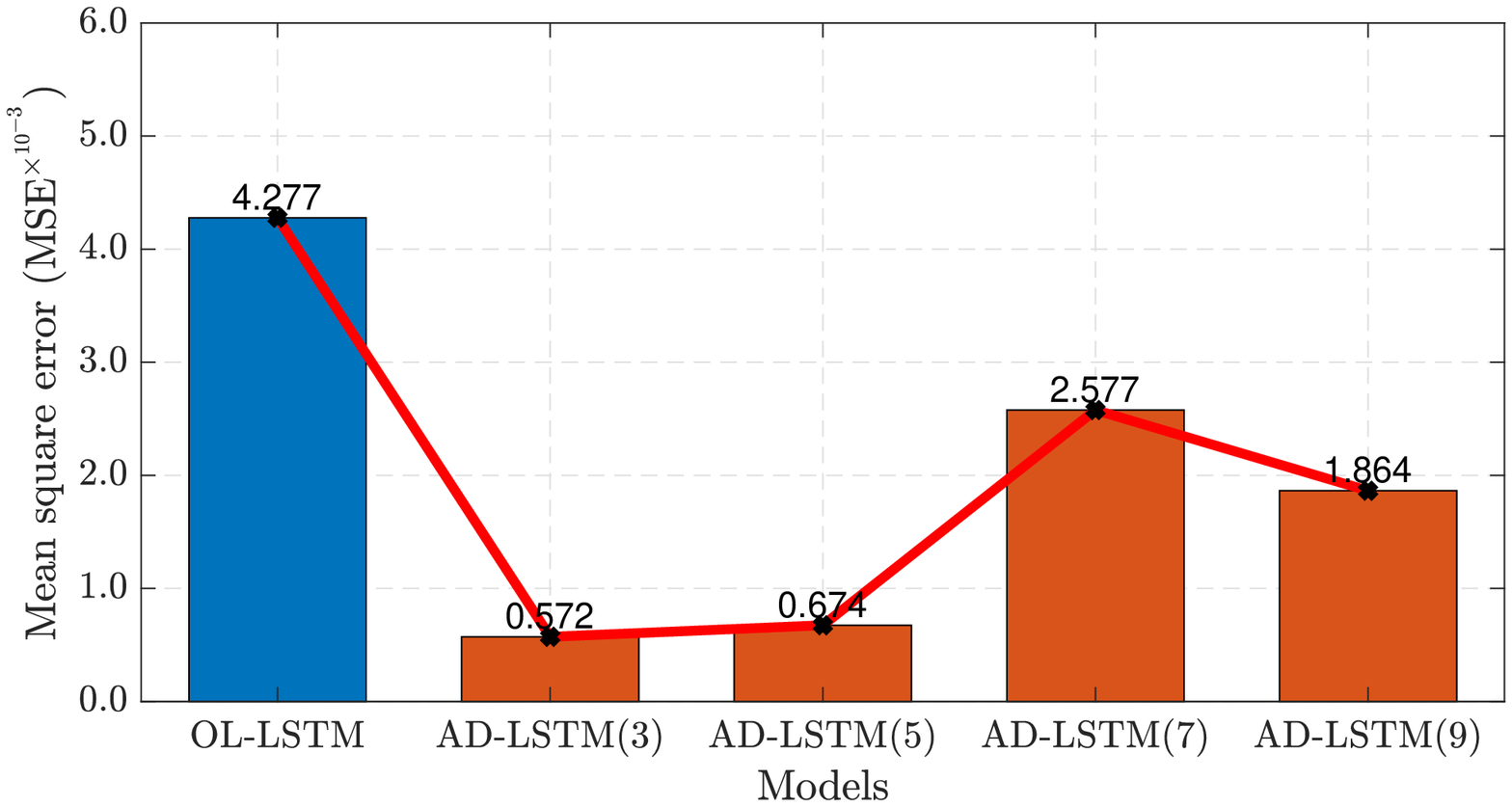}\vspace{-0.35in}
\subcaption{Case \#3 (Summer)}
\end{minipage} 
\begin{minipage}[t]{0.48\textwidth}\label{fig_7_Result_NoDrift_Autumn_d}
\centering
\includegraphics[width=\textwidth]{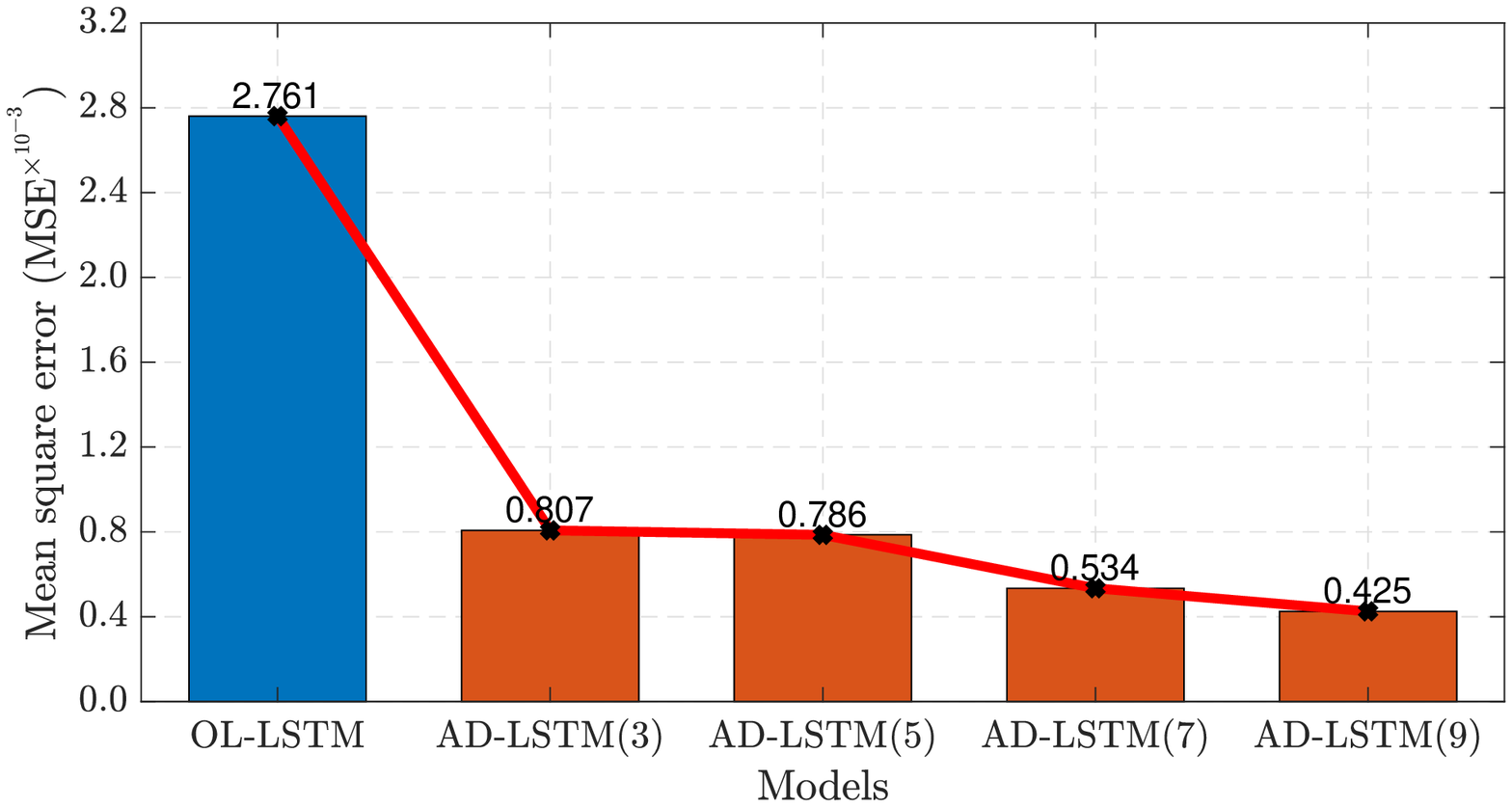}\vspace{-0.35in}
\subcaption{Case \#4 (Autumn)}
\end{minipage}\\ \vspace{-0.15in}
\caption{MSE results of day-ahead PVPG forecasting by OL-LSTM and AD-LSTM. Different sizes of training sets (i.e., training set = 3, 5, 7, and 9 days) are adopted. (Evaluated based on none-drift data). }
\label{fig:Result_NoDrift}
\end{figure}

\begin{figure}[H]
  \centering
  \includegraphics[width=14.5 cm]{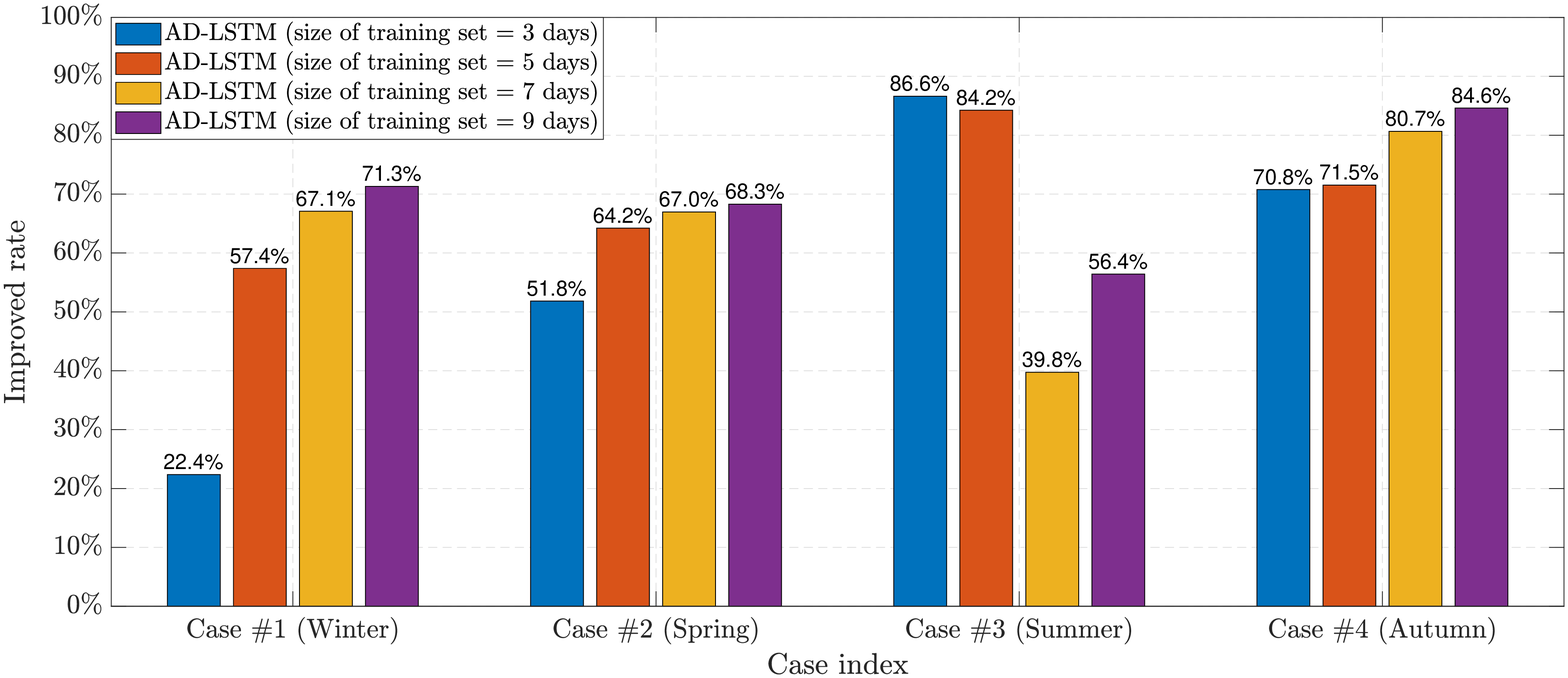}\vspace{-0.1in}
  \caption{Improved rate comparisons based on AD-LSTM with different sizes of training sets. (Evaluated based on none-drift data). }
  \label{fig:Improved_Rate_NoDrift}
\end{figure}

\begin{figure}[H]
\centering
\begin{minipage}[t]{0.48\textwidth}\label{fig_9_Result_NoDrift_Winter_a}
\centering
\includegraphics[width=\textwidth]{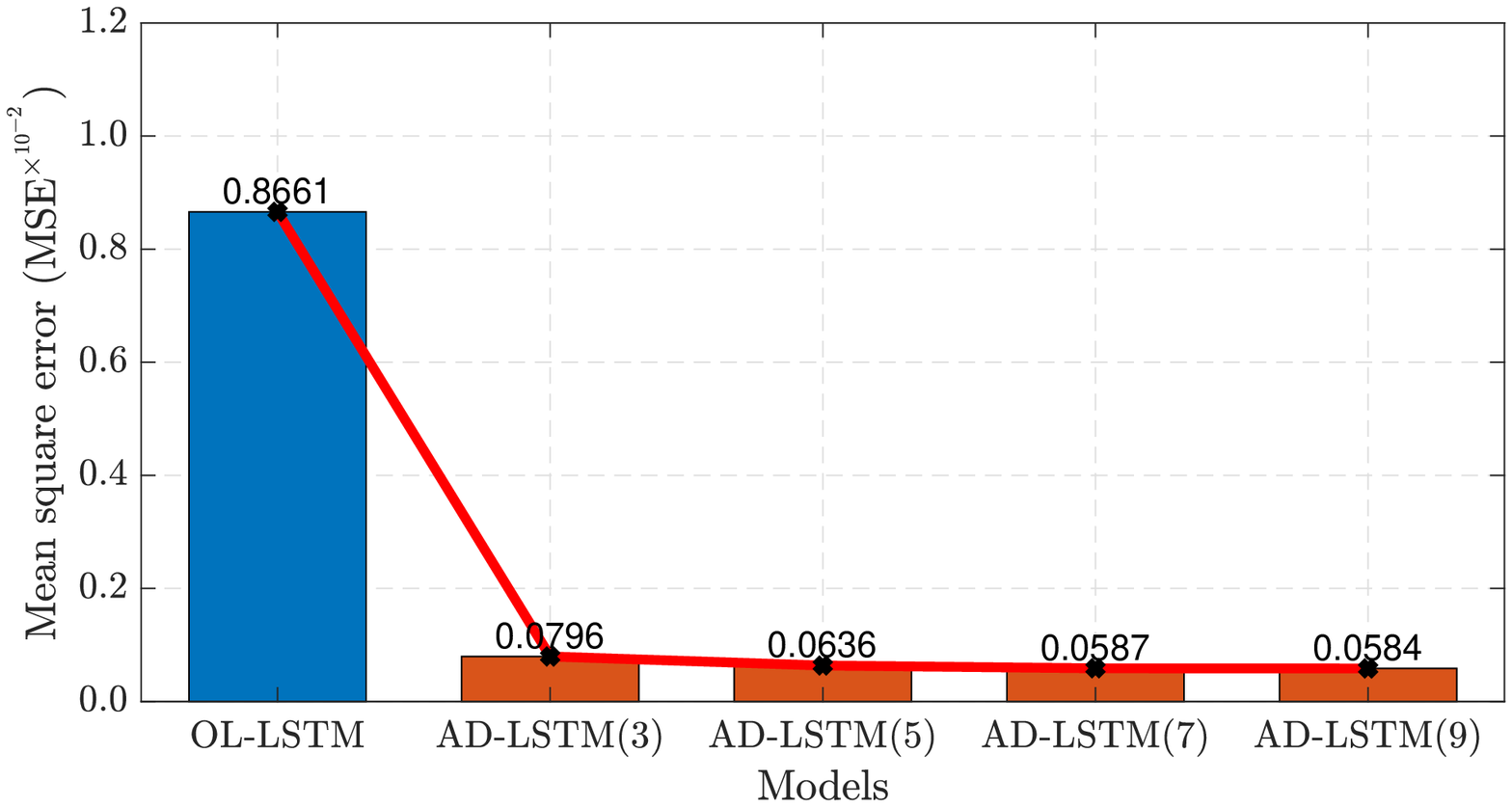}\vspace{-0.35in}
\subcaption{Case \#5 (Winter)}
\end{minipage} 
\begin{minipage}[t]{0.48\textwidth}\label{fig_9_Result_NoDrift_Spring_b}
\centering
\includegraphics[width=\textwidth]{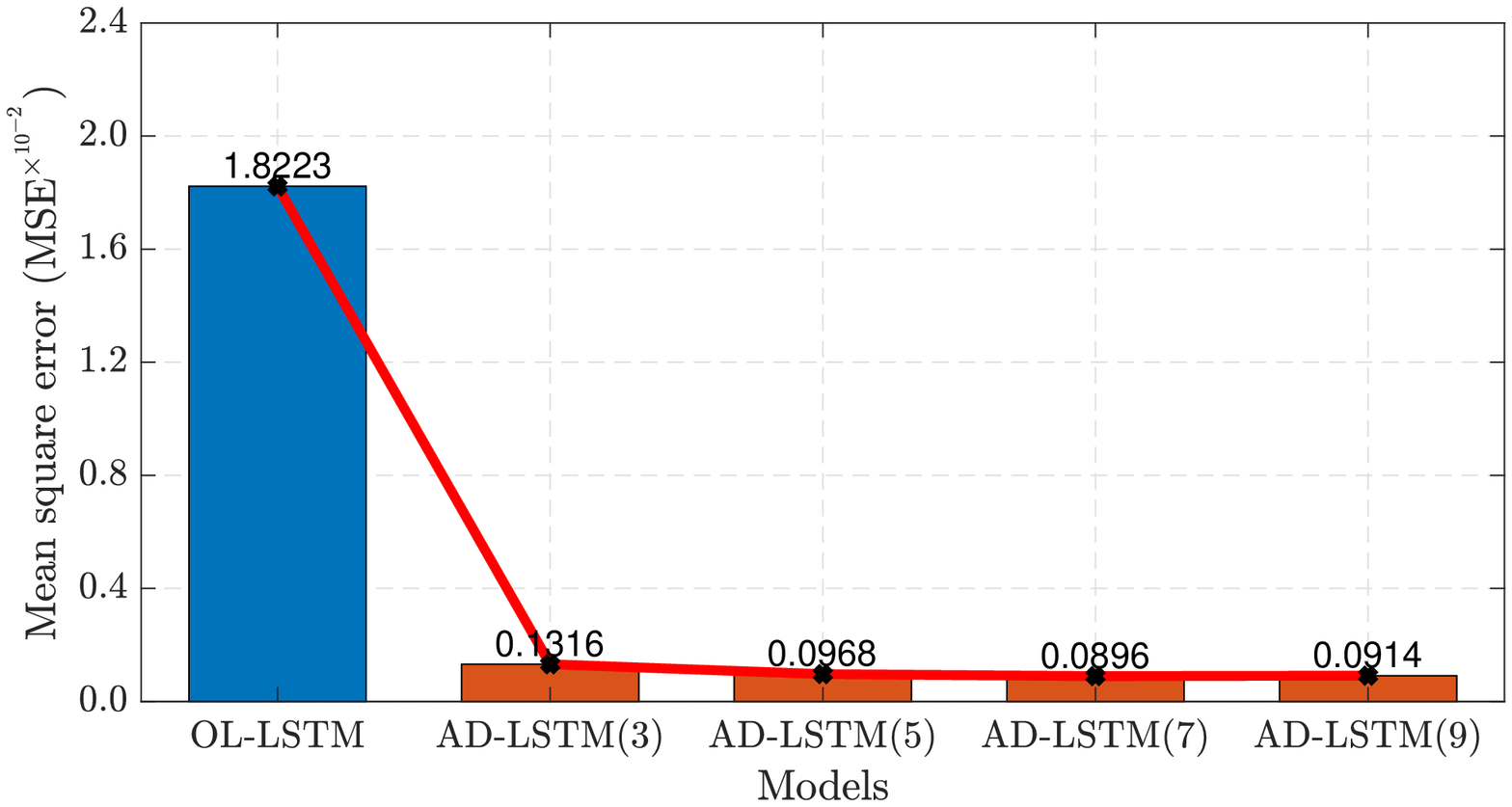}\vspace{-0.35in}
\subcaption{Case \#6 (Spring)}
\end{minipage}\\ 
\begin{minipage}[t]{0.48\textwidth}\label{fig_9_Result_NoDrift_Summer_c}
\centering
\includegraphics[width=\textwidth]{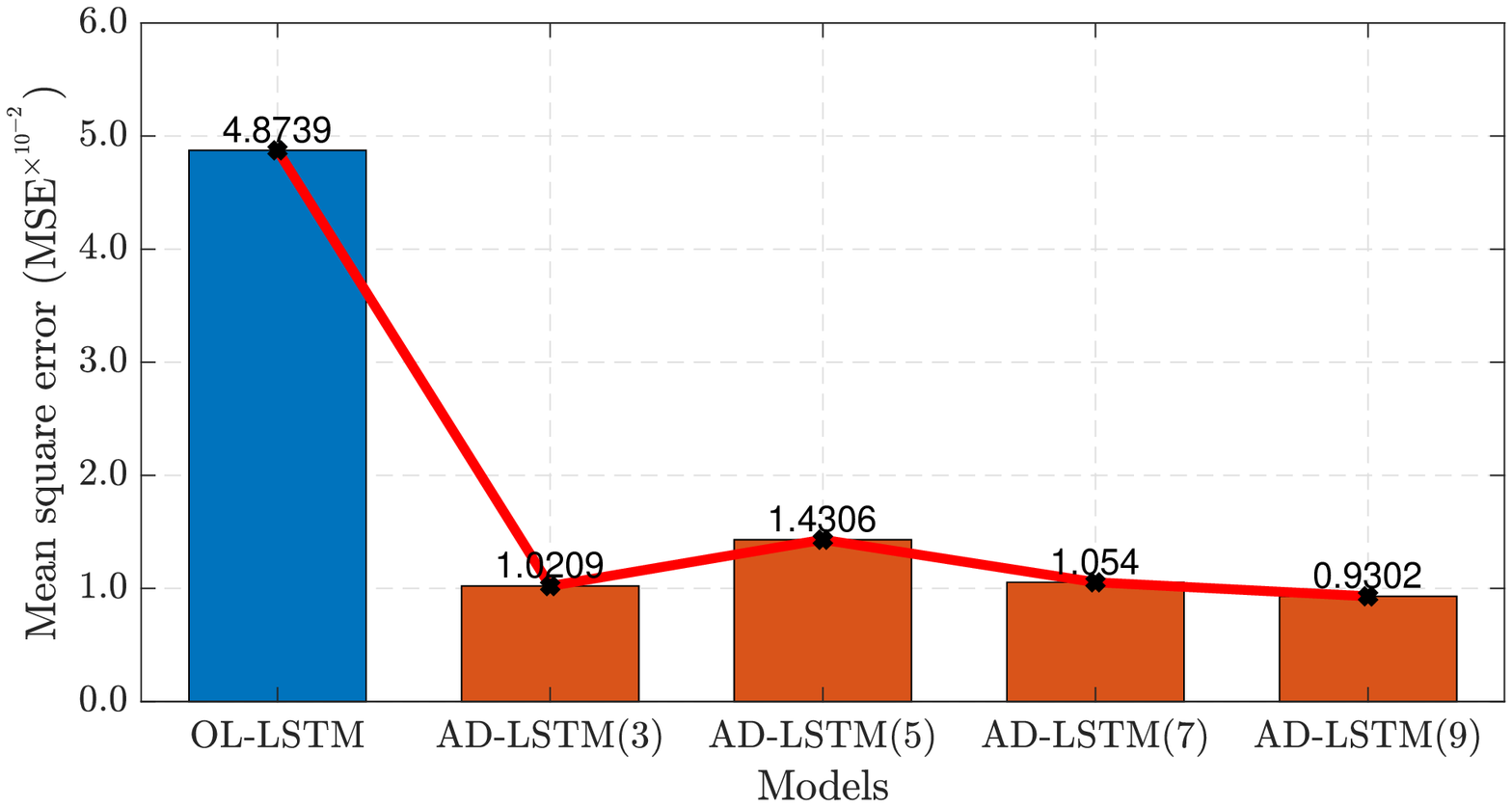}\vspace{-0.35in}
\subcaption{Case \#7 (Summer)}
\end{minipage} 
\begin{minipage}[t]{0.48\textwidth}\label{fig_9_Result_NoDrift_Autumn_d}
\centering
\includegraphics[width=\textwidth]{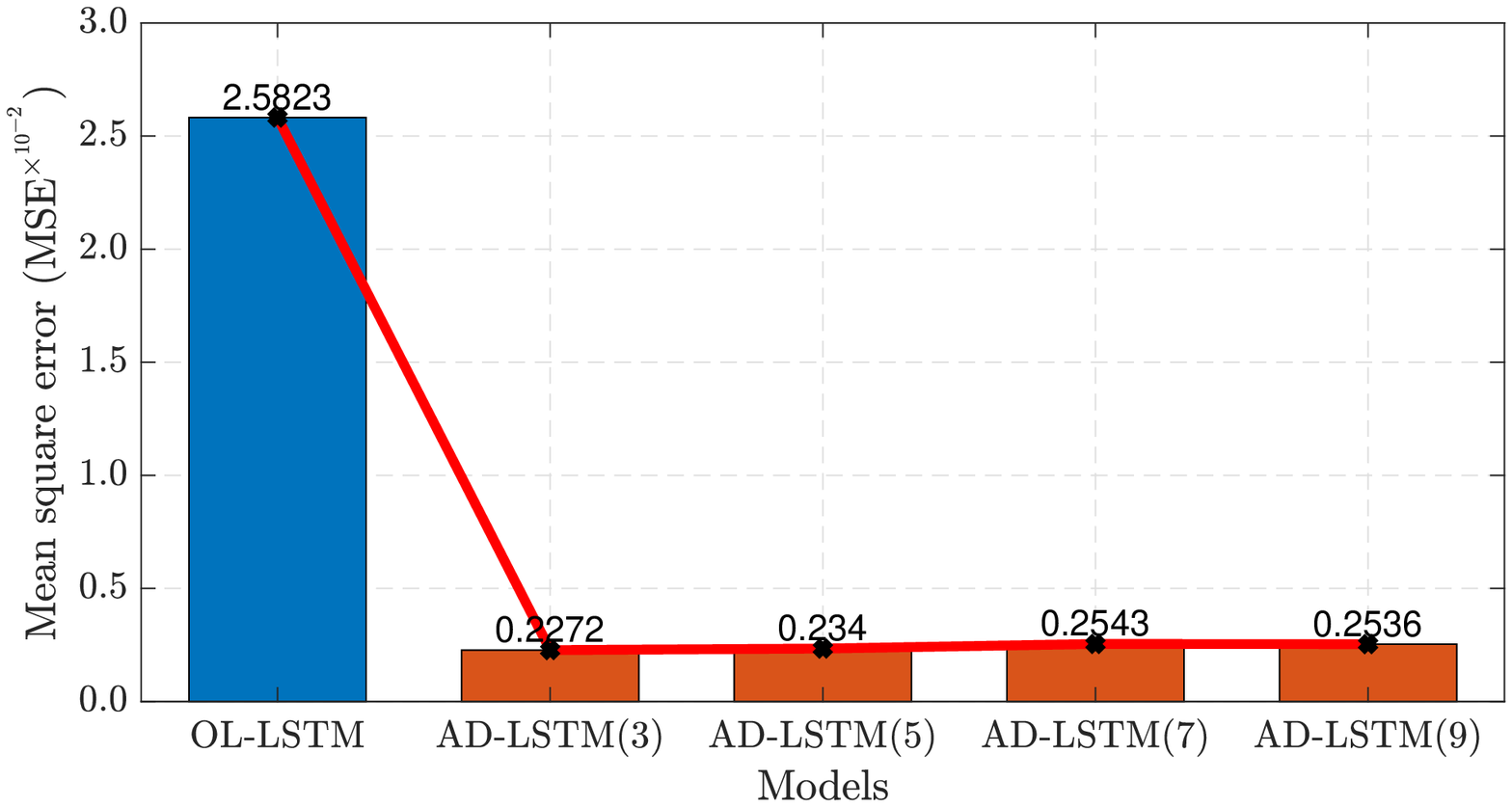}\vspace{-0.35in}
\subcaption{Case \#8 (Autumn)}
\end{minipage}\\ \vspace{-0.15in}
\caption{MSE results of day-ahead PVPG forecasting by OL-LSTM and AD-LSTM. Different sizes of training sets (i.e., training set = 3, 5, 7, and 9 days) are adopted. (Evaluated based on drift data). }
\label{fig:Result_Drift}
\end{figure}

\begin{figure}[H]
  \centering
  \includegraphics[width=14.5 cm]{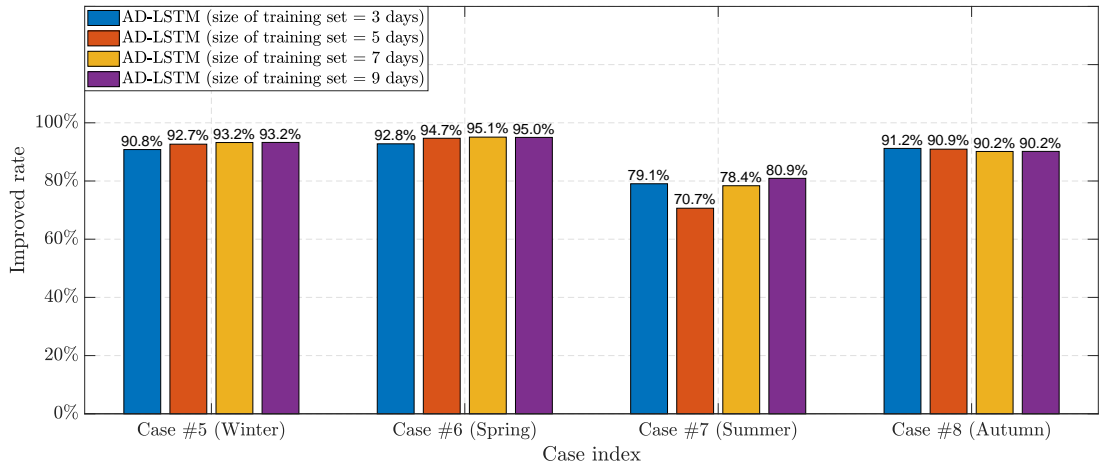}\vspace{-0.1in}
  \caption{Improved rate comparisons based on AD-LSTM with different sizes of training sets. (Evaluated based on drift data).}
  \label{fig:Improved_Rate_Drift}
\end{figure}

As the proposed AD-LSTM model is improved on the basis of the conventional offline LSTM (OL-LSTM) model by integrating the two-phase adaptive learning strategy (TP-ALS), it is necessary to compare the forecasting performances between AD-LSTM and OL-LSTM. Taking advantage of newly-arrived data, different sizes of supplementary training sets, i.e., size of training set = 3, 5, 7, and 9 days, are simultaneously evaluated. Therefore, the appropriate size of the training set can be determined. Case studies based on both none-drift and drift data are carried out in this subsection. To ensure more convincing results, as well as verifying the robustness of AD-LSTM, four individual cases from different seasons of a year are adopted in the evaluations. The MSE results of day-ahead PVPG forecasting by compared models for the none-drift data and the drift data are illustrated in Fig. \ref{fig:Result_NoDrift} and Fig. \ref{fig:Result_Drift}, respectively. The improved rate comparisons based on AD-LSTM with different sizes of training sets for the none-drift data and the drift data are presented in Fig. \ref{fig:Improved_Rate_NoDrift} and Fig. \ref{fig:Improved_Rate_Drift}, respectively. 

 \vspace{-0.2in}

 \vspace{-0.2in}

According to the results in Fig. \ref{fig:Result_NoDrift}, it is obvious that the AD-LSTM model achieves superior performance, with much lower MSE of PVPG forecasting, compared to the OL-LSTM model in all cases. The results meet the initial expectation since the AD-LSTM model not only takes advantage of historical data, but is also capable of continuously learning from the newly-arrived data. Essentially, utilizing the data which have higher correlations to the data of the target day as the supplementary dataset for model fine-tuning can make considerable contributions to accuracy improvements of the target day. In addition, it can also be observed that the AD-LSTM model with a bigger size of the training set (e.g., size of training set = 9 days) can produce better forecasts with a higher improved rate, as shown in Fig. \ref{fig:Improved_Rate_NoDrift}. The results conform to the characteristics of DL models, in which the more high-quality data that there are, the better is the performance of the model.

In terms of the evaluations based on the drift data, as shown in Fig. \ref{fig:Result_Drift}, the OL-LSTM model does not perform well in all cases due to the occurrences of concept drift. The OL-LSTM model cannot perform well on the data of which distributions have changed markedly. However, the proposed AD-LSTM model is still effective when concept drift appears, and can significantly improve forecasting accuracy compared to OL-LSTM. The results indicate that the proposed AD-LSTM model can adapt itself rapidly to capturing the newly-revealed patterns under streaming data. Additionally, it is interesting to find that the size of the training set may have limited impacts on accuracy improvements, as shown in Fig. \ref{fig:Improved_Rate_Drift}. The AD-LSTM model with a small size of training set (e.g., size of training set = 3 days) can also obtain an impressive accuracy improvement when compared to the model with bigger sizes of supplementary training sets. Since the proposed AD-LSTM does not need to re-train the old data, and the pre-trained model is fine-tuned based on the supplementary dataset which contains newly-arrived data on a small scale, it takes a very short time to prepare the AD-LSTM model before the target day.

\subsection{Day-ahead Forecasting of PV Power Generation}
\label{sec:55}

The performances of hourly day-ahead PVPG forecasting by different models are further compared in Subsection \ref{sec:55}. In this work, the proposed AD-LSTM model is mainly compared with a commonly-used persistence model (i.e., a forecasting model in which observations of the last day are persisted forward as the forecasting results) \cite{Xing-Luo-2021}, a statistical model (i.e., the auto-regressive integrated moving average model - ARIMA \cite{Vagro-2016}), a conventional machine learning model (i.e., the k-nearest neighbors - KNN \cite{Liu-Zhao-2016}), and the OL-LSTM model \cite{Jianqin-Zheng-2020, A-Gensler-2016, Abdel-Nasser-2017}. It is noted that the same hyper-parameters and network structures are adopted for both OL-LSTM and AD-LSTM. Based on the results in Subsection \ref{sec:54}, the sizes of the supplementary training sets are set as 9 days and 3 days for the none-drift data and the drift data, respectively.

\begin{figure}[H]
\centering
\begin{minipage}[t]{0.48\textwidth}\label{fig_11_Forecasting_Examples_a}
\centering
\includegraphics[width=\textwidth]{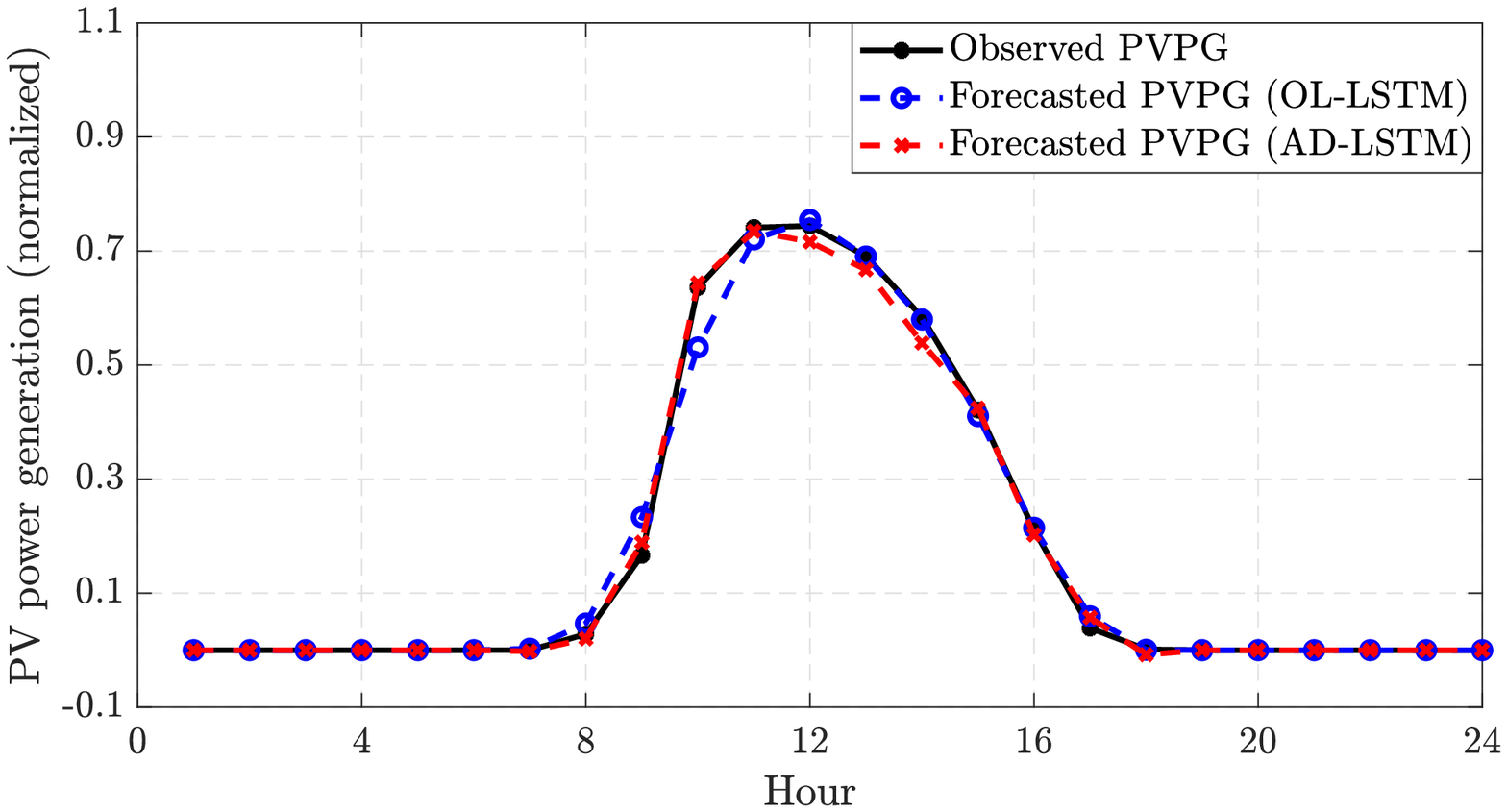}\vspace{-0.35in}
\subcaption{Case \#1 (Winter)}
\end{minipage} 
\begin{minipage}[t]{0.48\textwidth}\label{fig_11_Forecasting_Examples_a}
\centering
\includegraphics[width=\textwidth]{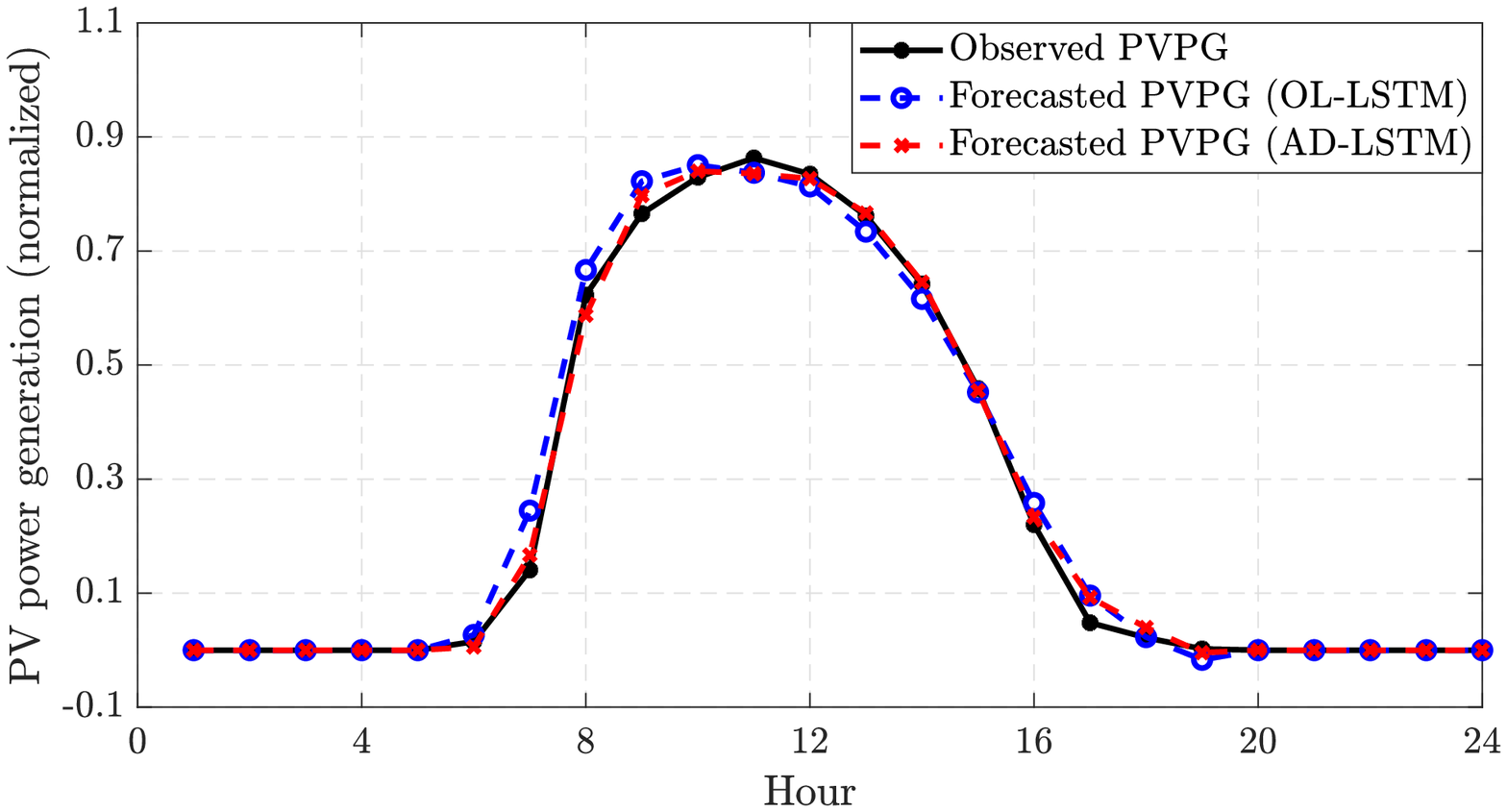}\vspace{-0.35in}
\subcaption{Case \#2 (Spring)}
\end{minipage} \\
\begin{minipage}[t]{0.48\textwidth}\label{fig_11_Forecasting_Examples_b}
\centering
\includegraphics[width=\textwidth]{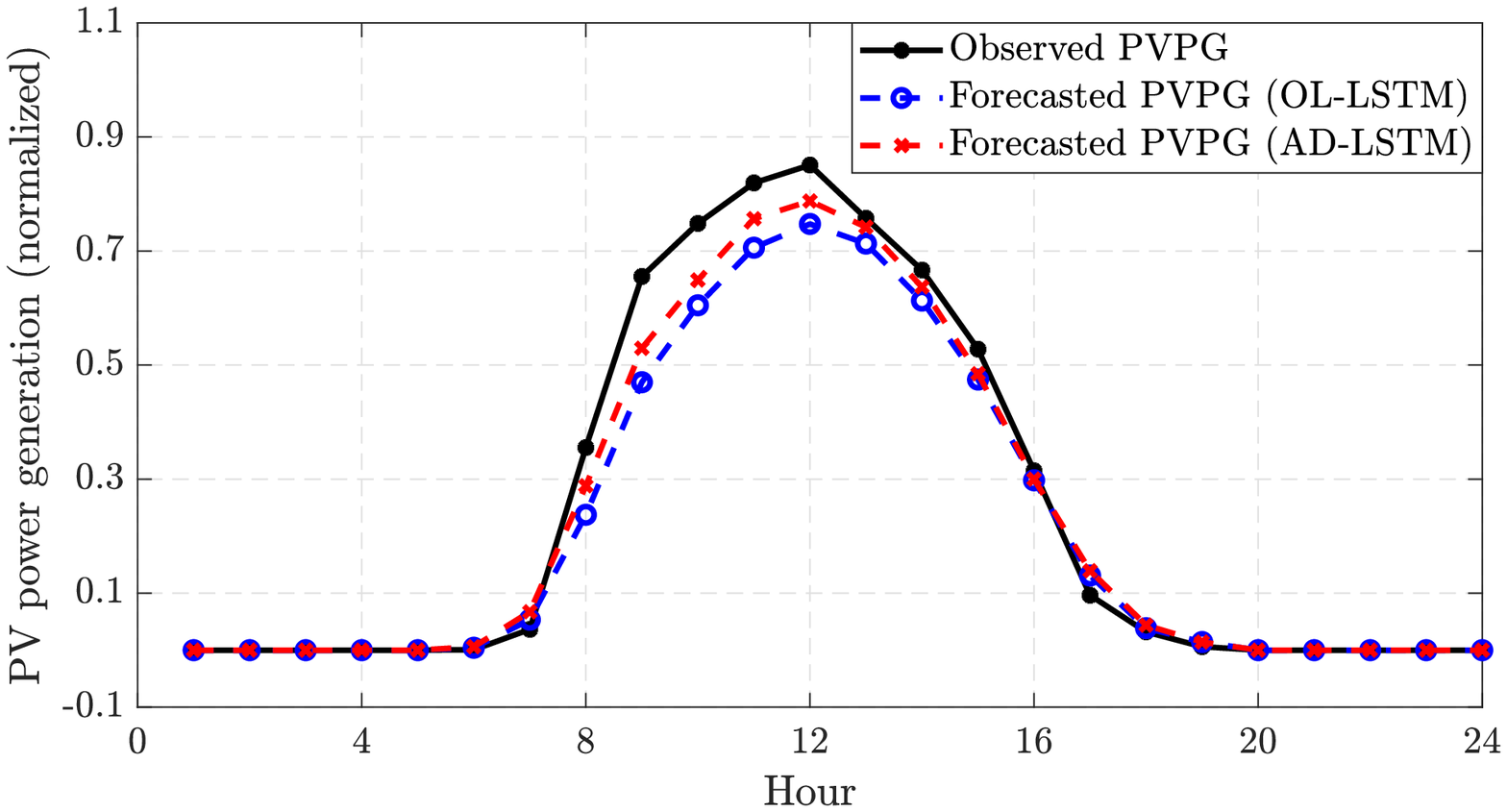}\vspace{-0.35in}
\subcaption{Case \#3 (Summer)}
\end{minipage} 
\begin{minipage}[t]{0.48\textwidth}\label{fig_11_Forecasting_Examples_c}
\centering
\includegraphics[width=\textwidth]{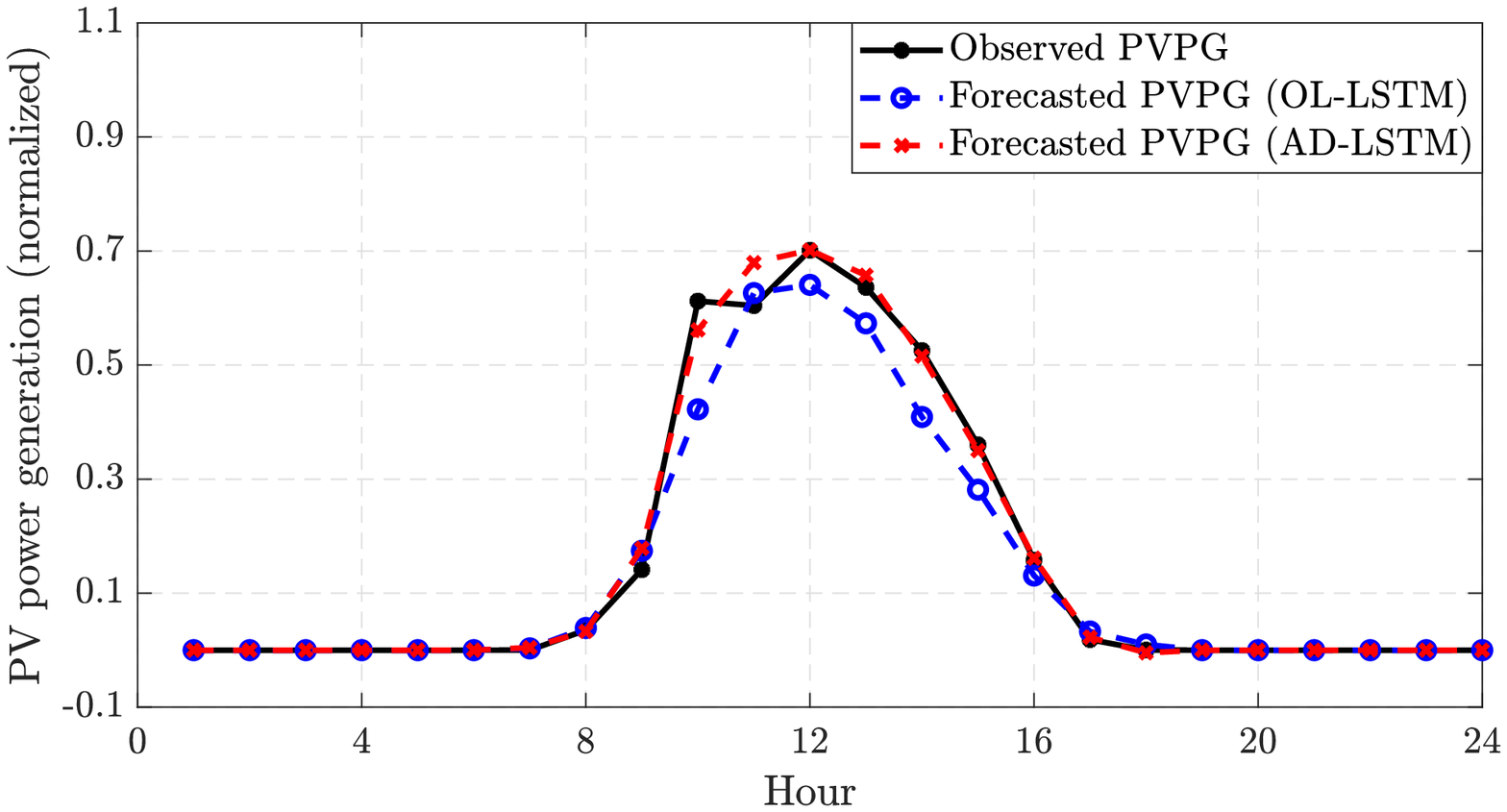}\vspace{-0.35in}
\subcaption{Case \#4 (Autumn)}
\end{minipage} \\
\begin{minipage}[t]{0.48\textwidth}\label{fig_11_Forecasting_Examples_d}
\centering
\includegraphics[width=\textwidth]{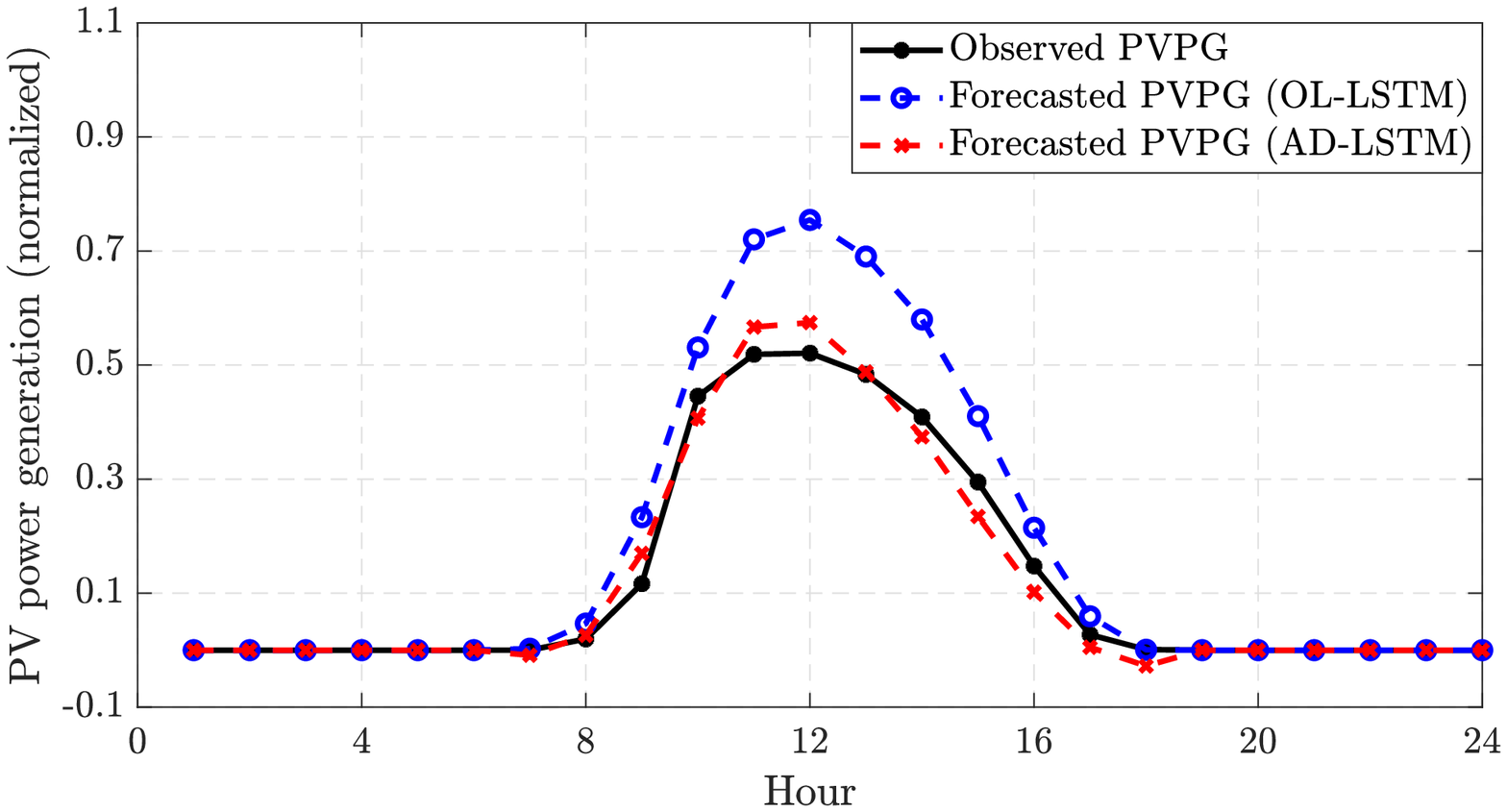}\vspace{-0.35in}
\subcaption{Case \#5 (Winter)}
\end{minipage} 
\begin{minipage}[t]{0.48\textwidth}\label{fig_11_Forecasting_Examples_e}
\centering
\includegraphics[width=\textwidth]{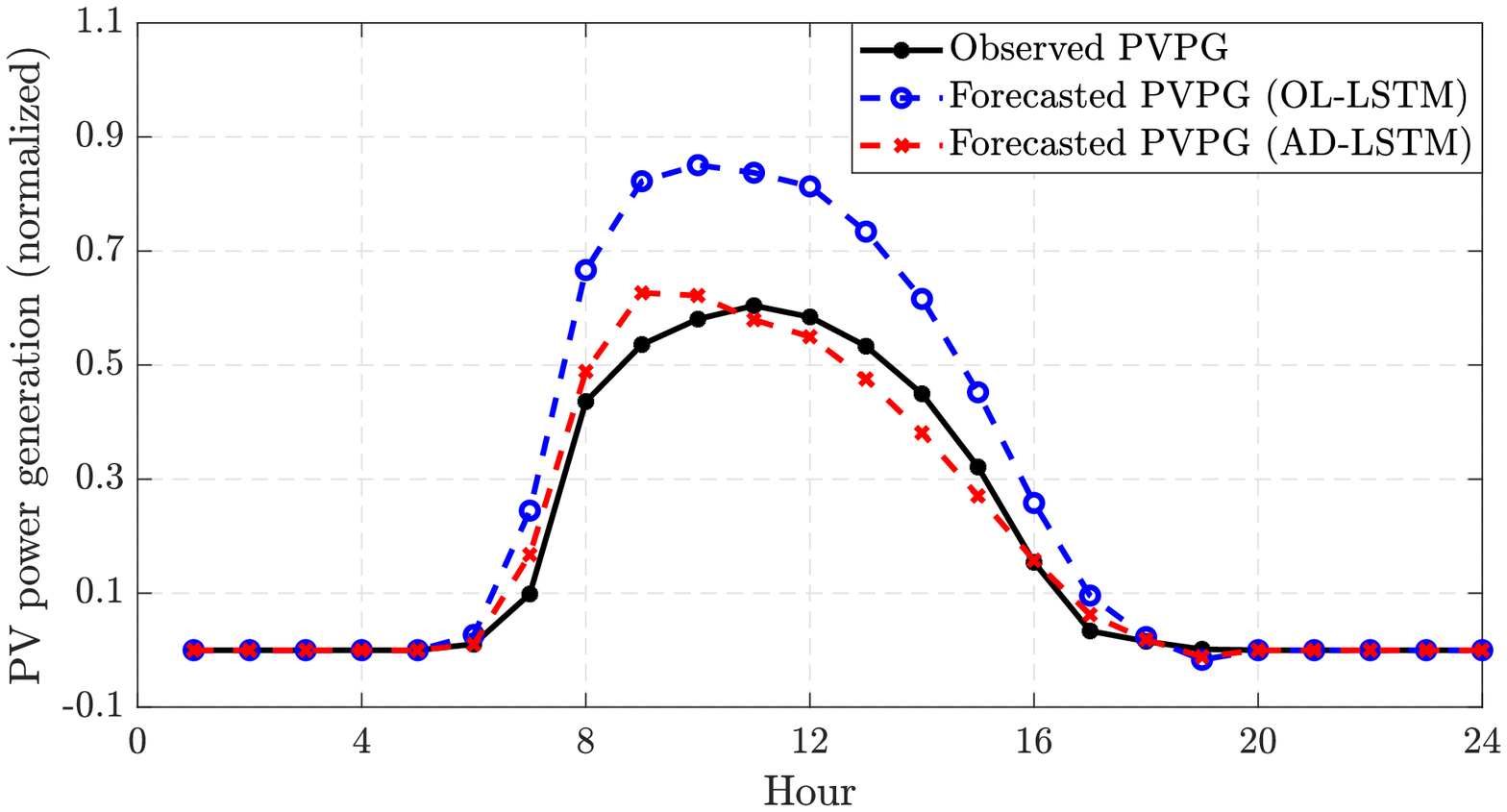}\vspace{-0.35in}
\subcaption{Case \#6 (Spring)}
\end{minipage} \\
\begin{minipage}[t]{0.48\textwidth}\label{fig_11_Forecasting_Examples_f}
\centering
\includegraphics[width=\textwidth]{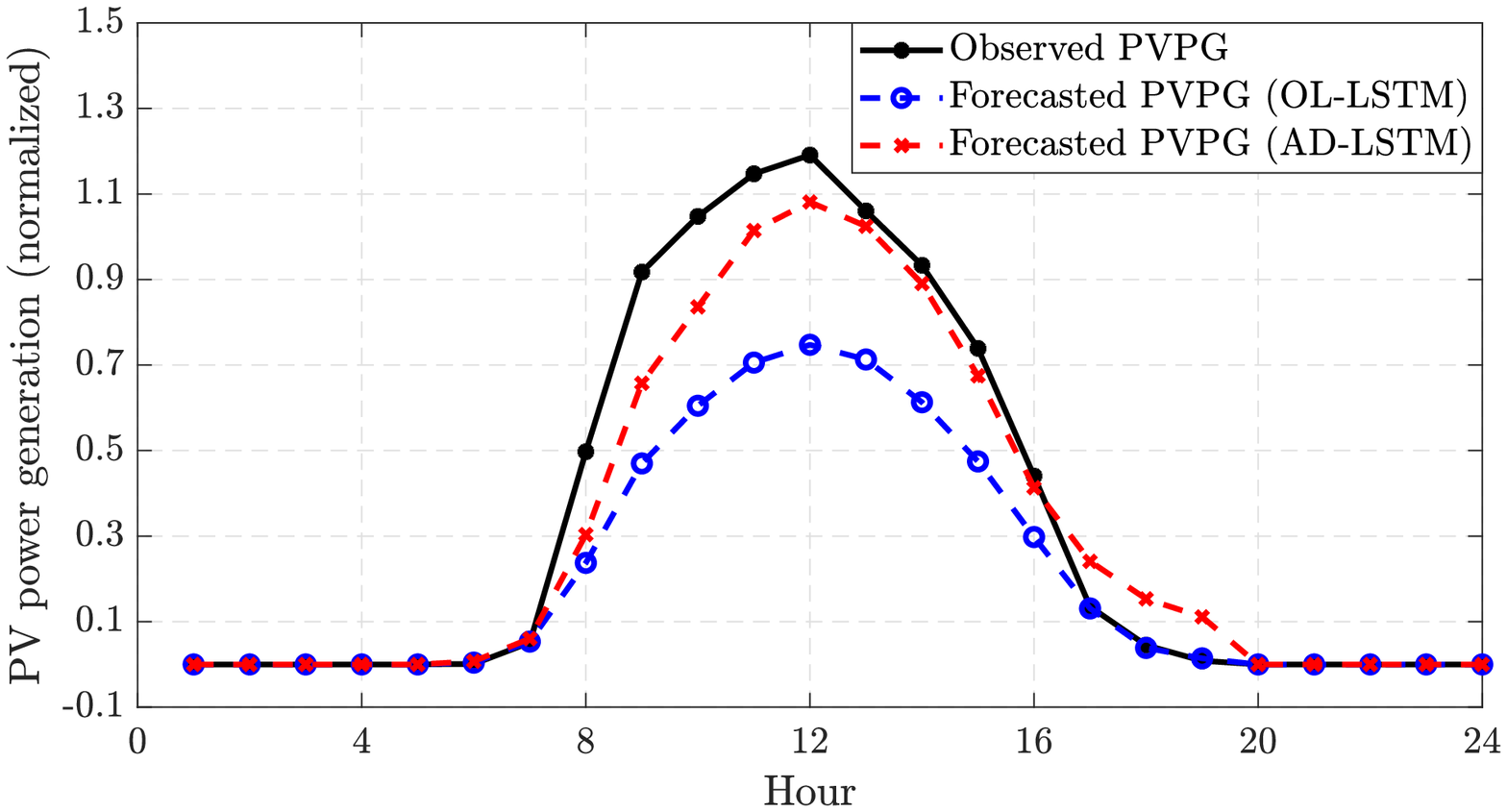}\vspace{-0.35in}
\subcaption{Case \#7 (Summer)}
\end{minipage} 
\begin{minipage}[t]{0.48\textwidth}\label{fig_11_Forecasting_Examples_a}
\centering
\includegraphics[width=\textwidth]{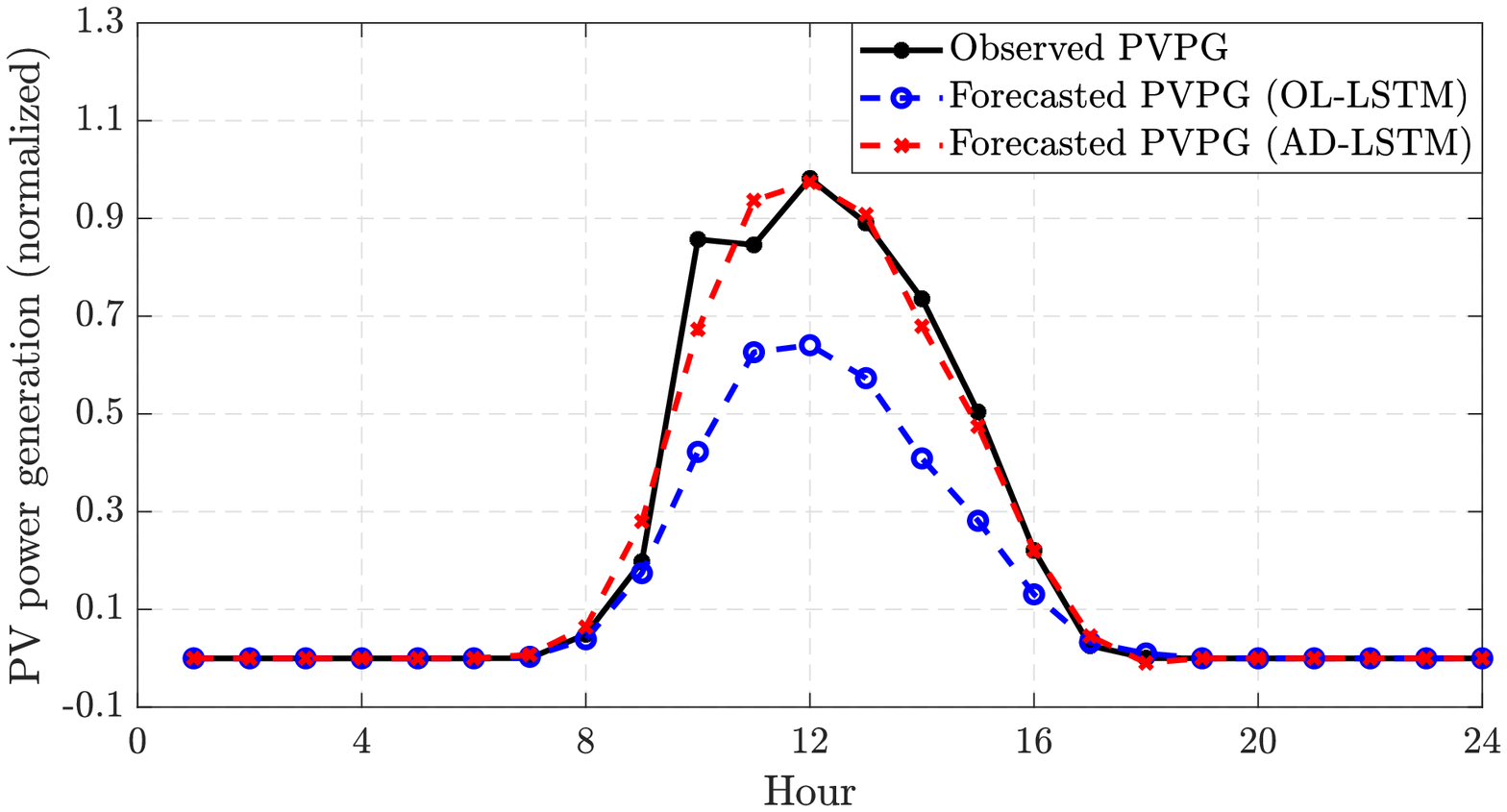}\vspace{-0.35in}
\subcaption{Case \#8 (Autumn)}
\end{minipage} \vspace{-0.15in}
\caption{Examples of visually displayed day-ahead PVPG forecasting results based on OL-LSTM and AD-LSTM. (Case \#1 - Case \#4 are evaluated based on the none-drift data, and Case \#5 - Case \#8 are evaluated based on the drift data).}
\label{fig:Forecasting_Examples}
\end{figure}

To better exhibit the day-ahead PVPG forecasting results based on the proposed DL models, examples of eight cases are presented in Fig. \ref{fig:Forecasting_Examples}. Case \#1 - Case \#4 are evaluated based on the none-drift data; whereas, Case \#5 - Case \#8 are evaluated based on the drift data. From the results in Fig. \ref{fig:Forecasting_Examples} (a) - (d), both OL-LSTM and AD-LSTM can fulfil the forecasting task with very high accuracy. The forecasted PVPG results follow the variations of the observed PVPG. Moreover, the proposed AD-LSTM model performs slightly better than the OL-LSTM model based on the obtained results. However, from the results in Fig. \ref{fig:Forecasting_Examples} (e) - (h), the OL-LSTM model demonstrates a weak PVPG forecasting capability and cannot capture the new patterns from the drift data. In contrast, the proposed AD-LSTM model can still maintain comparatively high accuracy of PVPG forecasting when concept drift appears.

The numerical results of PVPG forecasting comparisons between different models are illustrated in Table \ref{tab:Numerical_Results_Model_Comparison}.
Based on the results, it is apparent that the proposed AD-LSTM model performs the best among all compared models with lower values of MAE and MSE, and higher values of R$^{2}$ score. Specifically, DL models (i.e., OL-LSTM and AD-LSTM) outperform the persistence model, the ARIMA model, and the KNN model when comparing the results evaluated on the none-drift data. This indicates the superiority of DL models over the conventional statistics model and the machine learning model. The persistence model and the ARIMA model perform worse than other three compared models since they are only developed based on time-series PV data; however, meteorological data are not considered in model establishment. In addition, when evaluating the compared models based on the drift data, different results can be obtained. Specifically, the forecasting capabilities of KNN and OL-LSTM are significantly degraded, which shows that these models are not applicable for handling newly-arrived data with changes of data patterns. In contrast, the proposed AD-LSTM model maintains consistent performance in such evaluations, and demonstrates its superiority over other compared models. The results confirm the feasibility and effectiveness of the proposed AD-LSTM model in terms of improving the performance of day-ahead PVPG forecasting.

\begin{table}[H] 
\renewcommand\arraystretch{1.0}
\setlength{\abovecaptionskip}{0pt}
\setlength{\belowcaptionskip}{10pt}
\caption{Numerical results of PVPG forecasting comparisons between different forecasting models.}
\centering
\label{tab:Numerical_Results_Model_Comparison}\scalebox{0.68}{
\begin{tabular}{|l|l|cccc|cccc|}
\hline
\multirow{2}{*}{Model}    & \multirow{2}{*}{Eval.Criteria} & \multicolumn{4}{c|}{Evaluated based on none-drift data} & \multicolumn{4}{c|}{Evaluated based on drift data} \\ \cline{3-10}
                                   &                                      & Case \#1     & Case \#2     & Case \#3    & Case \#4    & Case \#5    & Case \#6   & Case \#7   & Case \#8   \\ \hline
\multirow{3}{*}{Persistence} & MAE$^{\times 10^{-2}}$               & 3.798        & 8.445        & 4.960       & 1.901       & 2.659       & 5.911      & 6.944      & 2.662      \\
                                   & MSE$^{\times 10^{-3}}$               & 7.373        & 36.978       & 11.290      & 2.395       & 3.613       & 18.119     & 22.128     & 4.694      \\
                                   & R$^2$ score                          & 0.903        & 0.679        & 0.891       & 0.961       & 0.903       & 0.679      & 0.891      & 0.961      \\ \hline
\multirow{3}{*}{ARIMA}       & MAE$^{\times 10^{-2}}$               & 3.960        & 5.072        & 5.500       & 5.030       & 2.773       & 3.422      & 6.819      & 7.062      \\
                                   & MSE$^{\times 10^{-3}}$               & 9.059        & 11.419       & 9.200       & 11.010      & 4.431       & 5.446      & 16.031     & 21.623     \\
                                   & R$^2$ score                          & 0.881        & 0.901        & 0.911       & 0.823       & 0.881       & 0.903      & 0.921      & 0.822      \\ \hline
\multirow{3}{*}{KNN}         & MAE$^{\times 10^{-2}}$               & 1.611        & 3.141        & 3.097       & 1.747       & 3.426       & 7.514      & 12.460     & 6.135      \\
                                   & MSE$^{\times 10^{-3}}$               & 1.032        & 3.667        & 4.662       & 1.464       & 4.445       & 13.030     & 43.793     & 13.933     \\
                                   & R$^2$ score                          & 0.986        & 0.968        & 0.955       & 0.976       & 0.880       & 0.769      & 0.784      & 0.885      \\ \hline
\multirow{3}{*}{OL-LSTM}     & MAE$^{\times 10^{-2}}$               & 1.187        & 2.078        & 3.993       & 2.796       & 5.332       & 8.960      & 13.286     & 8.564      \\
                                   & MSE$^{\times 10^{-3}}$               & 0.711        & 0.984        & 4.277       & 2.761       & 8.661       & 18.223     & 48.739     & 25.823     \\
                                   & R$^2$ score                          & 0.991        & 0.991        & 0.959       & 0.955       & 0.767       & 0.677      & 0.760      & 0.788      \\ \hline
\multirow{3}{*}{AD-LSTM}     & MAE$^{\times 10^{-2}}$               & \textbf{0.982}        & \textbf{1.388}        & \textbf{2.838}       & \textbf{1.068}       & \textbf{2.181}       & \textbf{2.580}      & \textbf{7.058}      & \textbf{2.457}      \\
                                   & MSE$^{\times 10^{-3}}$               & \textbf{0.203}        & \textbf{0.309}        & \textbf{1.879}       & \textbf{0.437}       & \textbf{0.797}       & \textbf{1.317}      & \textbf{9.979}      & \textbf{2.279}      \\
                                   & R$^2$ score                          & \textbf{0.997}        & \textbf{0.997}        & \textbf{0.982}       & \textbf{0.993}       & \textbf{0.979}       & \textbf{0.977}      & \textbf{0.951}      & \textbf{0.981}      \\ \hline
\end{tabular}}
\end{table}

\section{Conclusion}
\label{sec:6}

In this paper, we proposed an adaptive long short-term memory (AD-LSTM) model to improve day-ahead PVPG forecasting accuracy, as well as eliminate the impacts of concept drift. Incorporating with a two-phase adaptive learning strategy (TP-ALS), the AD-LSTM model can not only take advantage of historical data, but is also capable of continuously learning from newly-arrived data. In addition, a sliding window (SDWIN) algorithm was proposed to detect concept drift in PV systems. The obtained results indicated the stronger forecasting capability of AD-LSTM than the conventional OL-LSTM model, particularly in the presence of concept drift. The proposed AD-LSTM model also demonstrated superior performance in terms of day-ahead PVPG forecasting compared to other traditional machine learning models and statistical models in the literature.

In the future, the proposed approaches have the potential to be employed in other energy forecasting problems, including wind power generation forecasting, load demand forecasting, and real-time electricity price forecasting. Additionally, besides LSTM, the concept of adaptive learning may be combined with other emerging DL models, such as the gated recurrent unit (GRU) model \cite{W-Li-2019} and the transformer model \cite{Jacob-Devlin-2018}. Moreover, the proposed approaches can also be applied to other datasets collected from different geographical areas, such as the Middle East of North China, in future works.

\newpage
\section*{Appendix}
\setcounter{table}{0}
\renewcommand{\thetable}{A\arabic{table}}

Sensitive analysis of AD-LSTM with different structures is proposed in this Appendix. To observe the forecasting performance of AD-LSTM with different structures, numbers of hidden units in DL models are set as 4, 16, 64, 128, and 256, respectively. Except for the number of hidden units, the same hyper-parameters are adopted in each case to ensure fair comparisons. Identical to the setting in Subsection \ref{sec:55}, the sizes of the supplementary training sets are set as 9 days and 3 days for the none-drift data and the drift data, respectively. The training processes of DL models are executed on a high-performance CPU (type: Intel(R) Core(TM) i7-9700K CPU @ 3.60GHz), and the time costs for each case are also recorded. The forecasting performances of AD-LSTM with different model structures are illustrated in Table \ref{tab:Results of Different HS}. The best results of the comparative items produced by the DL models with different structures are marked in bold.

\begin{table}[H] 
\renewcommand\arraystretch{1.0}
\setlength{\abovecaptionskip}{0pt}
\setlength{\belowcaptionskip}{10pt}
\caption{Forecasting performances of AD-LSTM with different model structures.}
\centering
\label{tab:Results of Different HS}\scalebox{0.56}{
\begin{tabular}{|c|c|cccc|cccc|}
\hline
\multirow{2}{*}{Unit number} & \multirow{2}{*}{Comparative items} & \multicolumn{4}{c|}{Evaluated based on none-drift data} & \multicolumn{4}{c|}{Evaluated based on drift data} \\ \cline{3-10}
                                       &                                    & Case \#1      & Case \#2      & Case \#3     & Case \#4     & Case \#5     & Case \#6    & Case \#7    & Case \#8    \\ \hline
\multirow{4}{*}{Unit No. = 4}                  & MSE of OL-LSTM                            & $7.11\times10^{-4}$    & $\bm{9.84\times10^{-4}}$    & $4.28\times10^{-3}$   & $2.76\times10^{-3}$   & $8.66\times10^{-3}$   & $1.82\times10^{-2}$  & $4.87\times10^{-2}$  & $2.58\times10^{-2}$  \\
                                       & MSE of AD-LSTM                            & $1.99\times10^{-4}$    & $3.11\times10^{-4}$    & $1.77\times10^{-3}$   & $4.06\times10^{-4}$   & $7.92\times10^{-4}$   & $1.32\times10^{-3}$  & $\bm{9.82\times10^{-3}}$  & $2.28\times10^{-3}$  \\
                                       & MSE Imp.Rate                           & 72.0\%        & 68.4\%        & 58.7\%       & 85.3\%       & 90.9\%       & 92.7\%      & \textbf{79.8\%}      & 91.2\%      \\
                                       & Cost of AD-LSTM                               & \textbf{1084 sec}         & \textbf{1044 sec}         & \textbf{1036 sec}        & \textbf{1063 sec}        & \textbf{216 sec}         & \textbf{215 sec}        & 229 sec        & 228 sec       \\ \hline
\multirow{4}{*}{Unit No. = 16}                 & MSE of OL-LSTM                            & $4.43\times10^{-4}$    & $1.05\times10^{-3}$    & $4.63\times10^{-3}$   & $1.96\times10^{-3}$   & $8.91\times10^{-3}$   & $\bm{2.14\times10^{-3}}$  & $4.92\times10^{-2}$  & $2.36\times10^{-2}$  \\
                                       & MSE of AD-LSTMM                            & $9.80\times10^{-5}$    & $7.68\times10^{-4}$    & $1.81\times10^{-3}$   & $\bm{1.94\times10^{-4}}$   & $\bm{2.17\times10^{-4}}$   & $\bm{5.00\times10^{-4}}$  & $2.33\times10^{-2}$  & $2.04\times10^{-3}$  \\
                                       & MSE Imp.Rate                           & 77.9\%        & 26.6\%        & 60.9\%       & \textbf{90.1\%}       & \textbf{97.6\%}       & \textbf{97.7\%}      & 52.6\%      & 91.3\%      \\
                                       & Cost of AD-LSTM                               & 1088 sec        & 1063 sec        & 1066 sec       & 1066 sec       & 227 sec        & 231 sec       & \textbf{222 sec}       & \textbf{224 sec}       \\ \hline
\multirow{4}{*}{Unit No. = 64}                 & MSE of OL-LSTM                            & $5.14\times10^{-4}$    & $1.93\times10^{-3}$    & $\bm{3.73\times10^{-3}}$   & $1.94\times10^{-3}$   & $8.88\times10^{-3}$   & $2.27\times10^{-2}$  & $\bm{4.66\times10^{-2}}$  & $2.34\times10^{-2}$  \\
                                       & MSE of AD-LSTM                            & $1.33\times10^{-4}$    & $\bm{1.97\times10^{-4}}$    & $1.16\times10^{-3}$   & $3.94\times10^{-4}$   & $4.12\times10^{-4}$   & $8.29\times10^{-4}$  & $1.74\times10^{-2}$  & $2.12\times10^{-3}$  \\
                                       & MSE Imp.Rate                           & 74.1\%        & \textbf{89.8\%}        & 69.0\%       & 79.7\%       & 95.4\%       & 96.4\%      & 62.7\%      & 91.0\%      \\
                                       & Cost of AD-LSTM                               & 1216 sec        & 1228 sec        & 1246 sec       & 1240 sec       & 262 sec        & 277 sec       & 256 sec       & 261 sec       \\ \hline
\multirow{4}{*}{Unit No. = 128}                & MSE of OL-LSTM                            & $7.15\times10^{-4}$    & $1.06\times10^{-3}$    & $4.33\times10^{-3}$   & $\bm{1.78\times10^{-3}}$   & $9.22\times10^{-3}$   & $1.87\times10^{-2}$  & $4.91\times10^{-2}$  & $\bm{2.26\times10^{-2}}$  \\
                                       & MSE of AD-LSTM                            & $1.01\times10^{-4}$    & $4.87\times10^{-4}$    & $\bm{5.94\times10^{-4}}$   & $3.33\times10^{-4}$   & $4.65\times10^{-4}$   & $1.12\times10^{-3}$  & $1.57\times10^{-2}$  & $\bm{1.24\times10^{-3}}$  \\
                                       & MSE Imp.Rate                           & \textbf{85.9\%}        & 53.9\%        & \textbf{86.3\%}       & 81.3\%       & 95.0\%       & 94.0\%      & 68.0\%      & \textbf{94.5\%}      \\
                                       & Cost of AD-LSTM                               & 1666 sec        & 1664 sec        & 1666 sec       & 1678 sec       & 347 sec        & 350 sec       & 351 sec       & 350 sec       \\ \hline
\multirow{4}{*}{Unit No. = 256}                & MSE of OL-LSTM                            & $\bm{2.01\times10^{-4}}$    & $1.25\times10^{-3}$    & $4.11\times10^{-3}$   & $1.90\times10^{-3}$   & $\bm{8.02\times10^{-3}}$   & $1.83\times10^{-2}$  & $4.86\times10^{-2}$  & $2.40\times10^{-2}$  \\
                                       & MSE of AD-LSTM                            & $\bm{5.30\times10^{-5}}$    & $4.16\times10^{-4}$    & $8.22\times10^{-4}$   & $3.06\times10^{-4}$   & $3.85\times10^{-4}$   & $1.05\times10^{-3}$  & $1.04\times10^{-2}$  & $1.50\times10^{-3}$  \\
                                       & MSE Imp.Rate                           & 73.6\%        & 66.7\%        & 80.0\%       & 83.9\%       & 95.2\%       & 94.3\%      & 78.5\%      & 93.7\%      \\
                                       & Cost of AD-LSTM                               & 3716 sec        & 3707 sec        & 3590 sec       & 3769 sec       & 779 sec        & 764 sec       & 761 sec       & 783 sec       \\ \hline
\end{tabular}}
\end{table}

From the results in Table \ref{tab:Results of Different HS}, the following conclusions can be drawn: (\romannumeral1) Compared to the conventional OL-LSTM model, the AD-LSTM model with different structures can considerably improve day-ahead PVPG forecasting accuracy in each case, which is the same as the results in Subsection \ref{sec:54}. (\romannumeral2) When a more complex structure of the model is adopted, more computational resources are required. In general, the training time cost of AD-LSTM increases with the number of hidden units increasing. (\romannumeral3) A more complex structure of the DL model cannot guarantee better forecasting performance in experiments. The forecasting capability of a finalized model is also partially dependent on the selected pre-trained model obtained in the first phase of TP-ALS. The results in the Appendix are in line with the results in Section \ref{sec:5}, thus verifying the feasibility and effectiveness of the proposed AD-LSTM model.

\section*{Declaration of Competing Interests}

The authors declare that they have no known competing financial interests or personal relationships that could have appeared to influence the work reported in this paper.

\section*{Credit Authorship Contribution Statement}

\textbf{Xing Luo}: conceptualization, investigation, methodology, data curation, software, visualization, writing - original draft. \textbf{Dongxiao Zhang}: conceptualization, methodology, supervision, writing - review and editing, funding acquisition. 

\section*{Acknowledgements}

The research is partially supported with special funding from the Peng Cheng Laboratory and startup funding from the Southern University of Science and Technology (SUSTech).
%
%
%

\section*{References}
\bibliographystyle{elsarticle-num}
{\small
\bibliography{references}}

\begin{thebibliography}{10}
\expandafter\ifx\csname url\endcsname\relax
  \def\url#1{\texttt{#1}}\fi
\expandafter\ifx\csname urlprefix\endcsname\relax\def\urlprefix{URL }\fi
\expandafter\ifx\csname href\endcsname\relax
  \def\href#1#2{#2} \def\path#1{#1}\fi

\bibitem{Zhifeng-Guo-2018}
Z.~Guo, K.~Zhou, C.~Zhang, X.~Lu, W.~Chen, S.~Yang, Residential electricity
  consumption behavior: Influencing factors, related theories and intervention
  strategies, Renewable and Sustainable Energy Reviews 81 (2018) 399--412.
\newblock \href {https://doi.org/https://doi.org/10.1016/j.rser.2017.07.046}
  {\path{doi:https://doi.org/10.1016/j.rser.2017.07.046}}.

\bibitem{Mohammad-Navid-2021}
M.~N. Fekri, H.~Patel, K.~Grolinger, V.~Sharma, Deep learning for load
  forecasting with smart meter data: Online adaptive recurrent neural network,
  Applied Energy 282 (2021) 116177.
\newblock \href
  {https://doi.org/https://doi.org/10.1016/j.apenergy.2020.116177}
  {\path{doi:https://doi.org/10.1016/j.apenergy.2020.116177}}.

\bibitem{Zekai-Sen-2004}
Z.~Sen, Solar energy in progress and future research trends, Progress in Energy
  and Combustion Science 30~(4) (2004) 367--416.
\newblock \href {https://doi.org/https://doi.org/10.1016/j.pecs.2004.02.004}
  {\path{doi:https://doi.org/10.1016/j.pecs.2004.02.004}}.

\bibitem{C-A-Varotsos-2019}
C.~Varotsos, M.~Efstathiou, J.~Christodoulakis, Abrupt changes in global
  tropospheric temperature, Atmospheric Research 217 (2019) 114--119.
\newblock \href
  {https://doi.org/https://doi.org/10.1016/j.atmosres.2018.11.001}
  {\path{doi:https://doi.org/10.1016/j.atmosres.2018.11.001}}.

\bibitem{Xing-Luo-2021}
X.~Luo, D.~Zhang, X.~Zhu, Deep learning based forecasting of photovoltaic power
  generation by incorporating domain knowledge, Energy 225 (2021) 120240.
\newblock \href {https://doi.org/https://doi.org/10.1016/j.energy.2021.120240}
  {\path{doi:https://doi.org/10.1016/j.energy.2021.120240}}.

\bibitem{Rich-H-2013}
R.~H. Inman, H.~T. Pedro, C.~F. Coimbra, Solar forecasting methods for
  renewable energy integration, Progress in Energy and Combustion Science
  39~(6) (2013) 535--576.
\newblock \href {https://doi.org/https://doi.org/10.1016/j.pecs.2013.06.002}
  {\path{doi:https://doi.org/10.1016/j.pecs.2013.06.002}}.

\bibitem{M-N-Akhter-2019}
M.~N. {Akhter}, S.~{Mekhilef}, H.~{Mokhlis}, N.~{Mohamed Shah}, Review on
  forecasting of photovoltaic power generation based on machine learning and
  metaheuristic techniques, IET Renewable Power Generation 13~(7) (2019)
  1009--1023.
\newblock \href {https://doi.org/https://doi.org/10.1049/iet-rpg.2018.5649}
  {\path{doi:https://doi.org/10.1049/iet-rpg.2018.5649}}.

\bibitem{Matthew-Lave-2010}
M.~Lave, J.~Kleissl, Solar variability of four sites across the state of
  colorado, Renewable Energy 35~(12) (2010) 2867--2873.
\newblock \href {https://doi.org/https://doi.org/10.1016/j.renene.2010.05.013}
  {\path{doi:https://doi.org/10.1016/j.renene.2010.05.013}}.

\bibitem{Muhammad-Qamar-Raza-2016}
M.~Q. Raza, M.~Nadarajah, C.~Ekanayake, On recent advances in pv output power
  forecast, Solar Energy 136 (2016) 125--144.
\newblock \href {https://doi.org/https://doi.org/10.1016/j.solener.2016.06.073}
  {\path{doi:https://doi.org/10.1016/j.solener.2016.06.073}}.

\bibitem{Alessandro-Ferrara-2021}
A.~Ferrara, S.~Jakubek, C.~Hametner, Energy management of heavy-duty fuel cell
  vehicles in real-world driving scenarios: Robust design of strategies to
  maximize the hydrogen economy and system lifetime, Energy Conversion and
  Management 232 (2021) 113795.
\newblock \href
  {https://doi.org/https://doi.org/10.1016/j.enconman.2020.113795}
  {\path{doi:https://doi.org/10.1016/j.enconman.2020.113795}}.

\bibitem{Alberto-Dolara-2015}
A.~Dolara, S.~Leva, G.~Manzolini, Comparison of different physical models for
  pv power output prediction, Solar Energy 119 (2015) 83--99.
\newblock \href {https://doi.org/https://doi.org/10.1016/j.solener.2015.06.017}
  {\path{doi:https://doi.org/10.1016/j.solener.2015.06.017}}.

\bibitem{Daniel-Koster-2019}
D.~Koster, F.~Minette, C.~Braun, O.~O'Nagy, Short-term and regionalized
  photovoltaic power forecasting enhanced by reference systems on the example
  of luxembourg, Renewable Energy 132 (2019) 455--470.
\newblock \href {https://doi.org/https://doi.org/10.1016/j.renene.2018.08.005}
  {\path{doi:https://doi.org/10.1016/j.renene.2018.08.005}}.

\bibitem{Vagro-2016}
S.~I. Vagropoulos, G.~I. Chouliaras, E.~G. Kardakos, C.~K. Simoglou, A.~G.
  Bakirtzis, Comparison of sarimax, sarima, modified sarima and ann-based
  models for short-term pv generation forecasting, in: 2016 IEEE International
  Energy Conference (ENERGYCON), 2016, pp. 1--6.
\newblock \href {https://doi.org/10.1109/ENERGYCON.2016.7514029}
  {\path{doi:10.1109/ENERGYCON.2016.7514029}}.

\bibitem{John-Boland-2016}
J.~Boland, M.~David, P.~Lauret, Short term solar radiation forecasting: Island
  versus continental sites, Energy 113 (2016) 186--192.
\newblock \href {https://doi.org/https://doi.org/10.1016/j.energy.2016.06.139}
  {\path{doi:https://doi.org/10.1016/j.energy.2016.06.139}}.

\bibitem{X-Zhang-2019}
X.~{Zhang}, F.~{Fang}, J.~{Liu}, Weather-classification-mars-based photovoltaic
  power forecasting for energy imbalance market, IEEE Transactions on
  Industrial Electronics 66~(11) (2019) 8692--8702.
\newblock \href {https://doi.org/10.1109/TIE.2018.2889611}
  {\path{doi:10.1109/TIE.2018.2889611}}.

\bibitem{H-Sheng-2018}
H.~{Sheng}, J.~{Xiao}, Y.~{Cheng}, Q.~{Ni}, S.~{Wang}, Short-term solar power
  forecasting based on weighted gaussian process regression, IEEE Transactions
  on Industrial Electronics 65~(1) (2018) 300--308.
\newblock \href {https://doi.org/10.1109/TIE.2017.2714127}
  {\path{doi:10.1109/TIE.2017.2714127}}.

\bibitem{Liu-Zhao-2016}
Z.~Liu, Z.~Zhang, Solar forecasting by k-nearest neighbors method with weather
  classification and physical model, in: 2016 North American Power Symposium
  (NAPS), 2016, pp. 1--6.
\newblock \href {https://doi.org/10.1109/NAPS.2016.7747859}
  {\path{doi:10.1109/NAPS.2016.7747859}}.

\bibitem{John-Paul-Mueller-2016}
J.~P. Mueller, L.~Massaron, Machine Learning For Dummies, For Dummies, 2016.

\bibitem{M-Abuella-2015}
M.~{Abuella}, B.~{Chowdhury}, Solar power forecasting using artificial neural
  networks, in: North American Power Symposium, 2015, pp. 1--5.
\newblock \href {https://doi.org/10.1109/NAPS.2015.7335176}
  {\path{doi:10.1109/NAPS.2015.7335176}}.

\bibitem{J-Liu-2015}
J.~{Liu}, W.~{Fang}, X.~{Zhang}, C.~{Yang}, An improved photovoltaic power
  forecasting model with the assistance of aerosol index data, IEEE
  Transactions on Sustainable Energy 6~(2) (2015) 434--442.
\newblock \href {https://doi.org/10.1109/TSTE.2014.2381224}
  {\path{doi:10.1109/TSTE.2014.2381224}}.

\bibitem{Y-S-Manjili-2018}
Y.~S. {Manjili}, R.~{Vega}, M.~M. {Jamshidi}, Data-analytic-based adaptive
  solar energy forecasting framework, IEEE Systems Journal 12~(1) (2018)
  285--296.
\newblock \href {https://doi.org/10.1109/JSYST.2017.2769483}
  {\path{doi:10.1109/JSYST.2017.2769483}}.

\bibitem{Guido-Cervone-2017}
G.~Cervone, L.~C. Harding, S.~Alessandrini, L.~D. Monache, Short-term
  photovoltaic power forecasting using artificial neural networks and an analog
  ensemble, Renewable Energy 108 (2017) 274--286.
\newblock \href {https://doi.org/https://doi.org/10.1016/j.renene.2017.02.052}
  {\path{doi:https://doi.org/10.1016/j.renene.2017.02.052}}.

\bibitem{Abinet-Tesfaye-Eseye-2018}
A.~T. Eseye, J.~Zhang, D.~Zheng, Short-term photovoltaic solar power
  forecasting using a hybrid wavelet-pso-svm model based on scada and
  meteorological information, Renewable Energy 118 (2018) 357--367.
\newblock \href {https://doi.org/https://doi.org/10.1016/j.renene.2017.11.011}
  {\path{doi:https://doi.org/10.1016/j.renene.2017.11.011}}.

\bibitem{Fei-Wang-2019}
F.~Wang, Z.~Zhang, C.~Liu, Y.~Yu, S.~Pang, N.~Duic, M.~S. Khah, J.~P. Catalao,
  Generative adversarial networks and convolutional neural networks based
  weather classification model for day ahead short-term photovoltaic power
  forecasting, Energy Conversion and Management 181 (2019) 443--462.
\newblock \href
  {https://doi.org/https://doi.org/10.1016/j.enconman.2018.11.074}
  {\path{doi:https://doi.org/10.1016/j.enconman.2018.11.074}}.

\bibitem{Huaizhi-Wang-2019}
H.~Wang, Z.~Lei, X.~Zhang, B.~Zhou, J.~Peng, A review of deep learning for
  renewable energy forecasting, Energy Conversion and Management 198 (2019)
  111799.
\newblock \href
  {https://doi.org/https://doi.org/10.1016/j.enconman.2019.111799}
  {\path{doi:https://doi.org/10.1016/j.enconman.2019.111799}}.

\bibitem{Changsong-Chen-2011}
C.~Chen, S.~Duan, T.~Cai, B.~Liu, Online 24-h solar power forecasting based on
  weather type classification using artificial neural network, Solar Energy
  85~(11) (2011) 2856--2870.
\newblock \href {https://doi.org/https://doi.org/10.1016/j.solener.2011.08.027}
  {\path{doi:https://doi.org/10.1016/j.solener.2011.08.027}}.

\bibitem{Massaoudi-2019}
M.~Massaoudi, I.~Chihi, L.~Sidhom, M.~Trabelsi, S.~S. Refaat, F.~S. Oueslati,
  Performance evaluation of deep recurrent neural networks architectures:
  Application to pv power forecasting, in: 2019 2nd International Conference on
  Smart Grid and Renewable Energy (SGRE), 2019, pp. 1--6.
\newblock \href {https://doi.org/10.1109/SGRE46976.2019.9020965}
  {\path{doi:10.1109/SGRE46976.2019.9020965}}.

\bibitem{Wilms-2018}
H.~Wilms, M.~Cupelli, A.~Monti, On the necessity of exogenous variables for
  load, pv and wind day-ahead forecasts using recurrent neural networks, in:
  2018 IEEE Electrical Power and Energy Conference (EPEC), 2018, pp. 1--6.
\newblock \href {https://doi.org/10.1109/EPEC.2018.8598329}
  {\path{doi:10.1109/EPEC.2018.8598329}}.

\bibitem{Kusuma-2021}
V.~Kusuma, A.~Privadi, A.~L. Setya~Budi, V.~L. Budiharto~Putri, Photovoltaic
  power forecasting using recurrent neural network based on bayesian
  regularization algorithm, in: 2021 IEEE International Conference in Power
  Engineering Application (ICPEA), 2021, pp. 109--114.
\newblock \href {https://doi.org/10.1109/ICPEA51500.2021.9417833}
  {\path{doi:10.1109/ICPEA51500.2021.9417833}}.

\bibitem{Jianqin-Zheng-2020}
J.~Zheng, H.~Zhang, Y.~Dai, B.~Wang, T.~Zheng, Q.~Liao, Y.~Liang, F.~Zhang,
  X.~Song, Time series prediction for output of multi-region solar power
  plants, Applied Energy 257 (2020) 114001.
\newblock \href
  {https://doi.org/https://doi.org/10.1016/j.apenergy.2019.114001}
  {\path{doi:https://doi.org/10.1016/j.apenergy.2019.114001}}.

\bibitem{A-Gensler-2016}
A.~{Gensler}, J.~{Henze}, B.~{Sick}, N.~{Raabe}, Deep learning for solar power
  forecasting - an approach using autoencoder and lstm neural networks, in:
  IEEE International Conference on Systems, Man, and Cybernetics, 2016, pp.
  2858--2865.
\newblock \href {https://doi.org/10.1109/SMC.2016.7844673.}
  {\path{doi:10.1109/SMC.2016.7844673.}}

\bibitem{Abdel-Nasser-2017}
A.~N. Mohamed, M.~Karar, Accurate photovoltaic power forecasting models using
  deep lstm-rnn, Neural Computing and Applications (2017) 2727--2740.\href
  {https://doi.org/https://doi.org/10.1007/s00521-017-3225-z}
  {\path{doi:https://doi.org/10.1007/s00521-017-3225-z}}.

\bibitem{Lu-Jie-2019}
J.~Lu, A.~Liu, F.~Dong, F.~Gu, J.~Gama, G.~Zhang, Learning under concept drift:
  A review, IEEE Transactions on Knowledge and Data Engineering 31~(12) (2019)
  2346--2363.
\newblock \href {https://doi.org/10.1109/TKDE.2018.2876857}
  {\path{doi:10.1109/TKDE.2018.2876857}}.

\bibitem{M-Schuster-1997}
M.~{Schuster}, K.~K. {Paliwal}, Bidirectional recurrent neural networks, IEEE
  Transactions on Signal Processing 45~(11) (1997) 2673--2681.
\newblock \href {https://doi.org/10.1109/78.650093}
  {\path{doi:10.1109/78.650093}}.

\bibitem{Yao-Zhang-2016}
Y.~Zhang, J.~Wang, K-nearest neighbors and a kernel density estimator for
  gefcom2014 probabilistic wind power forecasting, International Journal of
  Forecasting 32~(3) (2016) 1074--1080.
\newblock \href
  {https://doi.org/https://doi.org/10.1016/j.ijforecast.2015.11.006}
  {\path{doi:https://doi.org/10.1016/j.ijforecast.2015.11.006}}.

\bibitem{Tao-Hong-2016}
T.~Hong, P.~Pinson, S.~Fan, H.~Zareipour, A.~Troccoli, R.~J. Hyndman,
  Probabilistic energy forecasting: Global energy forecasting competition 2014
  and beyond, International Journal of Forecasting 32~(3) (2016) 896--913.
\newblock \href
  {https://doi.org/https://doi.org/10.1016/j.ijforecast.2016.02.001}
  {\path{doi:https://doi.org/10.1016/j.ijforecast.2016.02.001}}.

\bibitem{W-Li-2019}
W.~{Li}, T.~{Logenthiran}, W.~L. {Woo}, Multi-gru prediction system for
  electricity generation's planning and operation, IET Generation, Transmission
  Distribution 13~(9) (2019) 1630--1637.
\newblock \href {https://doi.org/10.1049/iet-gtd.2018.6081}
  {\path{doi:10.1049/iet-gtd.2018.6081}}.

\bibitem{Jacob-Devlin-2018}
J.~Devlin, M.~Chang, K.~Lee, K.~Toutanova, Bert: Pre-training of deep
  bidirectional transformers for language understanding (2018).
\newblock \href {http://arxiv.org/abs/1810.04805} {\path{arXiv:1810.04805}}.

\end{thebibliography}

\end{document}